%% file: main.tex
\documentclass[10pt,twocolumn,letterpaper]{article}

\usepackage[pagenumbers]{cvpr} 

\usepackage{times}
\usepackage{epsfig}
\usepackage{graphicx}
\usepackage{amsmath}
\usepackage{amssymb}
\usepackage{enumitem}

\input{00_custom.tex}

\usepackage[pagebackref=true,breaklinks=true,letterpaper=true,colorlinks,bookmarks=false]{hyperref}

\usepackage[capitalize]{cleveref}
\crefname{section}{Sec.}{Secs.}
\Crefname{section}{Section}{Sections}
\Crefname{table}{Table}{Tables}
\crefname{table}{Tab.}{Tabs.}

\def\cvprPaperID{6639} 
\def\confName{CVPR}
\def\confYear{2022}

\begin{document}

\title{\methodname: Fast 3D Pose from Multiple Views without 3D Supervision}

\author{Ben Usman$^{1,2}$\\
{\tt\small usmn@bu.edu}
\and
Andrea Tagliasacchi$^{2,3}$\\
{\tt\small taglia@google.com}
\and
Kate Saenko$^{1,4}$\\
{\tt\small saenko@bu.edu}
\and
Avneesh Sud$^{2}$\\
{\tt\small asud@google.com}
\and
$^{1}$Boston University, \ \ \ $^{2}$Google Research, \ \ \ $^{3}$University of Toronto, \ \ \ $^{4}$MIT-IBM Watson AI Lab
}

\maketitle

\input{0_abstract}
\input{1_introduction}
\input{2_related}
\input{3_method}
\input{4_experiments}
\input{5_results}

\input{6_conclusions}
\input{8_ack}

\clearpage
{
    \small
    \bibliographystyle{plainnat}
    \bibliography{main}
}
\clearpage

\input{7_supplementary}

\end{document}

%% file: 00_custom.tex
\usepackage[export]{adjustbox}
\usepackage{booktabs}
\usepackage{multirow}
\usepackage{graphicx}
\usepackage{bm}

\usepackage[dvipsnames]{xcolor}
\usepackage[numbers]{natbib}
\usepackage{stackengine}
\usepackage{afterpage}

\usepackage{amssymb}
\usepackage{pifont}
\newcommand{\cmark}{\ding{51}}%
\newcommand{\xmark}{\ding{55}}%

\usepackage{algorithm}
\usepackage{algpseudocode}
\usepackage{algorithmicx}
\algrenewcommand\algorithmicrequire{\textbf{Input:}}
\algrenewcommand\algorithmicensure{\textbf{Output:}}

\usepackage{colortbl}
\usepackage{forloop}

\newcommand*\diff{\mathop{}\!\mathrm{d}}
\newcommand*\Prob{p}

\newcommand{\inctabcolsep}[2]{\addtolength{\tabcolsep}{#1} #2 \addtolength{\tabcolsep}{-#1}}

\newcommand*\methodname{MetaPose}

\newcounter{rowcntr}[table]

\newcommand{\CIRCLE}[1]{\raisebox{.5pt}{\footnotesize \textcircled{\scriptsize \raisebox{-0.7pt}{#1}}}}

\newcommand{\Figure}[1]{Figure~\ref{fig:#1}}

\newcommand{\Table}[1]{Table~\ref{tab:#1}}
\newcommand{\eq}[1]{\eqref{eq:#1}}

\newcommand{\Section}[1]{Section~\ref{sec:#1}}

\renewcommand{\paragraph}[1]{\vspace{.25em}\noindent\textbf{#1}.}

\usepackage{blindtext}

\usepackage{colortbl}
\usepackage{arydshln,graphicx,xcolor,array}

\DeclareMathOperator*{\argmin}{arg\,min}

\usepackage{enumitem}
\usepackage{diagbox}

%% file: 0_abstract.tex
\begin{abstract}
In the era of deep learning, human pose estimation from multiple cameras with unknown calibration has received little attention to date.
We show how to train a neural model to perform this task with high precision and minimal latency overhead. The proposed model takes into account joint location uncertainty due to occlusion from multiple views, and requires only 2D keypoint data for training.
Our method outperforms both classical bundle adjustment and weakly-supervised monocular 3D baselines on the well-established Human3.6M dataset, as well as the more challenging in-the-wild Ski-Pose PTZ dataset.
\end{abstract}

%% file: 1_introduction.tex
\section{Introduction}

We tackle the problem of estimating 3D coordinates of human joints from RGB images captured using synchronized (potentially moving) cameras with unknown positions, orientations, and intrinsic parameters. We additionally assume having access to a training set with \textit{only} 2D positions of joints labeled on captured images.

Historically, real-time capture of the human 3D pose has been undertaken only by large enterprises that could afford expensive specialized motion capture equipment~\cite{gleicher1999animation}.
In principle, if camera calibrations are available~\cite{building_rome_10.1145/2001269.2001293}, human body joints can be triangulated directly from camera-space observations~\cite{iskakov2019learnable, karashchuk2020anipose}.
One scenario in which camera calibration cannot easily be estimated is sports capture, in which close-ups of players are captured in front of \textit{low-texture backgrounds}, with \textit{wide-baseline, moving cameras}.
Plain backgrounds preclude calibration, as not sufficiently many feature correspondences can be detected across views; see~\Figure{teaser}.

\input{figs/teaser}
In this work, we propose a neural network to simultaneously predict 3D human and relative camera poses from multiple views; see~\Figure{teaser}.
Our approach uses \textit{human body} joints as a source of information for camera calibration.
As joints often become occluded, \textit{uncertainty} must be carefully accounted for, to avoid bad calibration and consequent erroneous 3D pose predictions.
As we assume a synchronized multi-camera setup at test-time, our algorithm should also be able to effectively \textit{aggregate} information from different viewpoints.
Finally, our approach supervised by 2D annotations \textit{alone},  as ground-truth annotation of 3D data is unwieldy.
As summarized in \Figure{priorwork}, and detailed in what follows, none of the existing approaches fully satisfies these fundamental requirements.

\noindent\textbf{Fully-supervised 3D} pose estimation approaches yield the lowest estimation error, but make use of known 3D camera specification during either training \cite{xie2020metafuse} or both training and inference \cite{iskakov2019learnable}. However, the prohibitively high cost of 3D joint annotation and full camera calibration in-the-wild makes it difficult to acquire large enough labeled datasets representative of specific environments~\cite{rhodin2018skipose, joo2020exemplar}, therefore rendering supervised methods not applicable in this setup.

\noindent\textbf{Monocular 3D} methods, \cite{iqbal2020weakly, kocabas2019epipolar, wandt2020canonpose} as well as 2D-to-3D lifting networks \cite{chen2019unsupervised, wandt2019repnet}, relax data constraints to enable 3D pose inference using just multi-view 2D data without calibration at train time.
Unfortunately, at inference time, these methods can only be applied to a single view at a time, therefore unable to leverage cross-view information and uncertainty.

\noindent\textbf{Classical SfM} (structure from motion) approaches to 3D pose estimation \cite{karashchuk2020anipose} iteratively refine both the camera and the 3D pose from noisy 2D observations.
However, these methods are often much slower than their neural counterparts, since they have to perform several optimization steps during inference.
Further, most of them do not consider uncertainty estimates, resulting in sub-par performance.

\input{figs/priorwork}

To overcome these limitation we propose~\textit{\methodname}; see~\Figure{teaser}.
Our method for 3D pose estimation aggregates pose predictions and uncertainty estimates across multiple views, requires no 3D joint annotations or camera parameters at both train and inference time, and adds very little latency to the resulting pipeline.

Overall, we propose the feed-forward neural architecture that can accurately estimate the 3D human pose \textit{and} the relative cameras configuration from multiple views, taking into account joint occlusions and prediction uncertainties, and uses only 2D joint annotations for training.
We employ an off-the-shelf weakly-supervised 3D network to form an \textit{initial} guess about the pose and the camera setup, and a neural \textit{meta}-optimizer that iteratively \textit{refines} this guess using 2D joint location probability heatmaps generated by an off-the-shelf 2D pose estimation network.
This modular approach not only yields low estimation error, leading to state-of-the-art results on Human3.6M~\cite{ionescu2013human36m} and Ski-Pose PTZ~\cite{rhodin2018skipose}, but also has low latency, as inference within our framework executes as a feed-forward neural network.


%% file: figs/teaser.tex
\begin{figure}[t]
\begin{center}
\includegraphics[width=\linewidth]{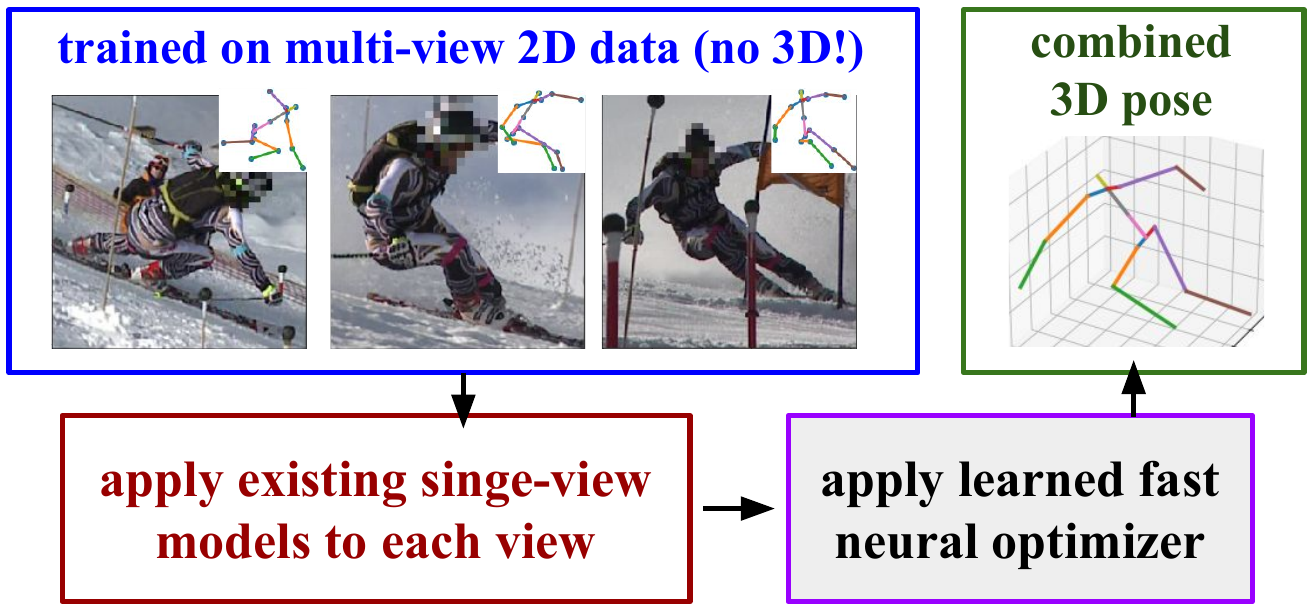}
\end{center}
\vspace{-1.5em}
\caption{
We show how to train a neural network that can aggregate outputs of multiple single-view methods, takes prediction uncertainty into consideration, has minimal latency overhead, and requires only 2D supervision for training.
Our method mimics the structure of bundle-adjustment solvers, but using the joints of the human body to drive camera calibration, and by implementing a bundle-like solver with a simple feed-forward neural network.\vspace{-5px}
} %
\label{fig:teaser}
\end{figure}

%% file: figs/priorwork.tex
\begin{figure}[t]
\begin{center}
\includegraphics[width=\linewidth,trim=0 3.3in 4.6in 0,clip]{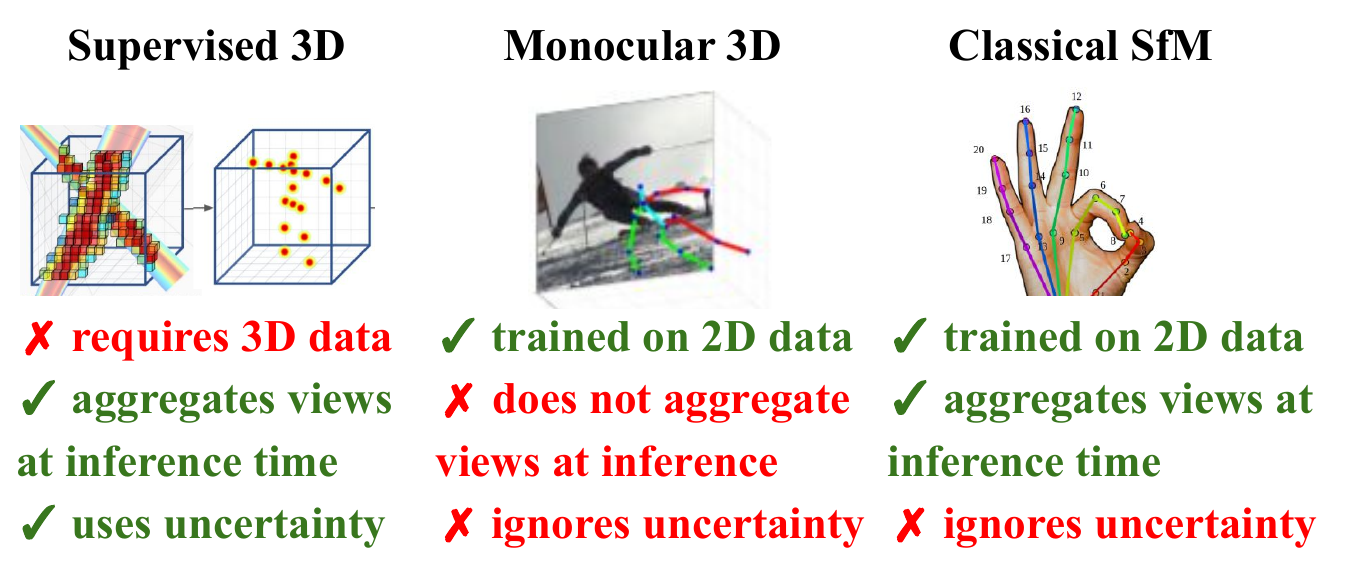}
\end{center}
\vspace{-1.5em}
\caption{
\textit{Prior work} ---
Existing solutions either 
require 3D annotations \cite{iskakov2019learnable}, 
perform inference on a single view at a time \cite{wandt2020canonpose}, or 
ignore uncertainty in joint coordinates due to occlusions \cite{karashchuk2020anipose}. 
%
}
\label{fig:priorwork}
\end{figure}

%% file: 2_related.tex
\section{Related Work}

In this section, we review \textit{multi-view} 3D human pose estimation, while we refer to~\citet{joo2020exemplar} for a survey of 3D human pose estimation in the wild.

\paragraph{Full supervision}
Supervised methods \cite{iskakov2019learnable, tu2020voxelpose, Chen_2020_CVPR_crossview} yield the lowest 3D pose estimation errors on multi-view single person~\cite{ionescu2013human36m} and multi-person~\cite{Joo_2019_panoptic, belagiannis20143d, Chen_2020_CVPR_crossview} datasets, but require precise camera calibration during both training and inference. Other approaches \cite{xie2020metafuse} use datasets with full 3D annotations and a large number of annotated cameras to train methods that can adapt to novel camera setups in visually similar environments, therefore somewhat relaxing camera calibration requirements.
\citet{martinez2017simple} use pre-trained 2D pose networks \cite{newell2016stacked} to take advantage of existing datasets with 2D pose annotations. 
Epipolar transformers~\cite{epipolartransformers} use only 2D keypoint supervision, but require camera calibration to incorporate 3D information in the 2D feature extractors.

\input{figs/method2d}

\paragraph{Weak and self-supervision} 
Several approaches exist that do not use paired 3D ground truth data. Many augment limited 3D annotations with 2D labels~\cite{Zhou_2017_weakly_supervised, hmrKanazawa17, Mitra_2020_CVPR, zanfir2020weakly}. Fitting-based methods~\cite{hmrKanazawa17, kocabas2020vibe, kolotouros2019spin, zanfir2020weakly} jointly fit a statistical 3D human body model and 3D human pose to monocular images or videos. Analysis-by-synthesis methods~\cite{Rhodin_2018_ECCV, Kundu_2020_nvs, Jakab_2020_CVPR_unlabelled} learn to predict 3D human pose by estimating appearance in a novel view. 
Most related to our work are approaches that exploit the structure of multi-view image capture.
EpipolarPose~\cite{kocabas2019epipolar} uses epipolar geometry to obtain 3D pose estimates from multi-view 2D predictions, and subsequently uses them to directly supervise 3D pose regression.
\citet{iqbal2020weakly} proposes a weakly-supervised baseline to predict pixel coordinates of joints and their depth in each view and penalized the discrepancy between rigidly aligned predictions for different views during training.
The self-supervised CanonPose~\cite{wandt2020canonpose} further advances state-of-the-art by decoupling 3D pose estimation in ``canonical'' frame.
\citet{drover2018can} learn a ``dictionary'' mapping 2D pose projections into corresponding realistic 3D poses, using a large collection of simulated 3D-to-2D projections. RepNet~\cite{wandt2019repnet} and~\citet{chen2019unsupervised} train similar ``2D-to-3D lifting networks'' with more realistic data constraints.
While all the aforementioned methods use multi-view consistency for \textit{training}, they do not allow pose \textit{inference} from multiple images.

\paragraph{Iterative refinement}
Estimating camera and pose simultaneously is a long-standing problem in vision~\cite{rosales2001estimating}.
One of the more recent successful attempts is the work of~\citet{bridgeman2019multi} that proposed an end-to-end network that refines the initial calibration guess using center points of multiple players in the field.
In the absence of such external calibration signals, \citet{takahashi2018human} performs bundle adjustment with bone length constraints,  
but do not report results on a public benchmark.
AniPose~\cite{karashchuk2020anipose} performs joint 3D pose and camera refinement using a modified version of the robust 3D registration algorithm of~\citet{zhou2016fast}. Such methods ignore predicted uncertainty for faster inference, but robustly iteratively estimate outlier 2D observations and ignores them during refinement.
In Section~\ref{sec:results}, we show that these classical approaches struggle in ill-defined settings, such as when we have a small number of cameras. More recently, SPIN~\cite{kolotouros2019spin}, HUND~\cite{zanfir2021neural} and Holopose~\cite{Guler_2019_CVPR_holopose} incorporate iterative pose refinement for \textit{monocular} inputs, however, the refinement is tightly integrated into the pose estimation network. \methodname~effectively regularizes the \textit{multi-view} pose estimation problem with a finite-capacity neural network resulting in both faster inference and higher precision than the classical refinement.

%% file: figs/method2d.tex
\begin{figure*}
\centering
\includegraphics[width=\linewidth]{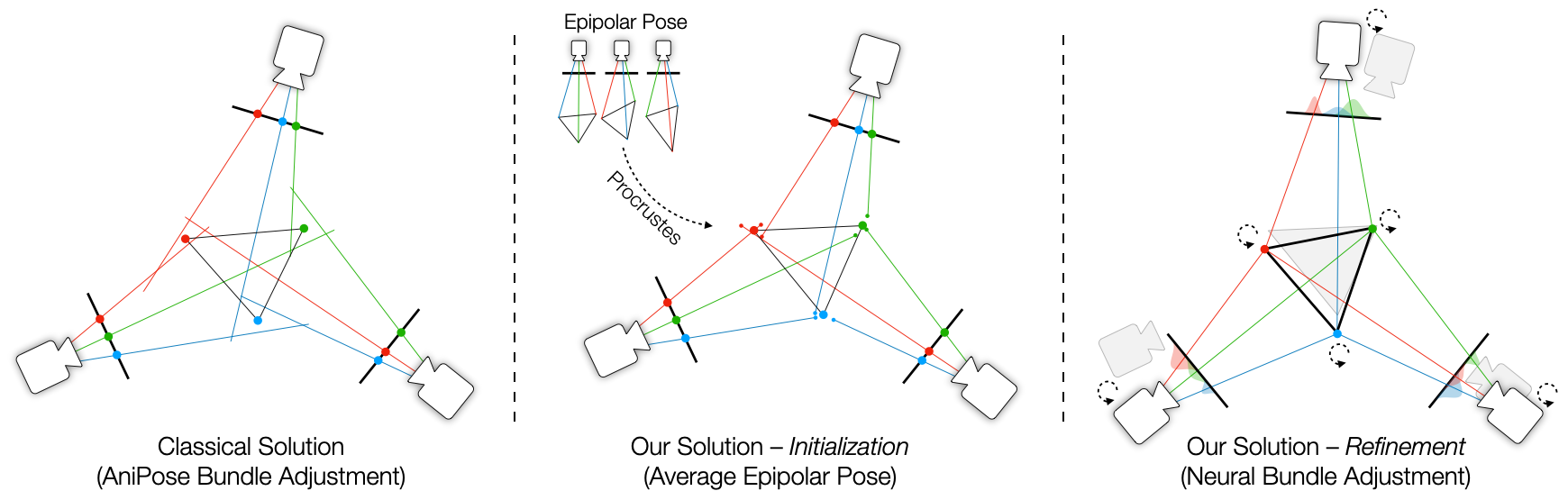}
\vspace{-10px}
\caption{
\textbf{Method} --
We illustrate our method with a simple 2D example of regressing the 3D vertices of an equilateral triangle given multi-view observations.
(left) AniPose~\cite{karashchuk2020anipose} performs classical bundle adjustment to identify camera positions and 3D vertices that minimize reprojection error to 2D landmarks on the input images.
Conversely, our technique \textit{emulates} classical bundle adjustment in a ``neural'' fashion by a meta-optimizer:
first (middle), the EpipolarPose~\cite{kocabas2019epipolar} neural network obtains a per-frame 3D estimate of the joints, which we co-align via procrustes to estimate an initialization of both 3D cameras and 3D joints; 
then (right), a neural network meta-optimizer performs a bundle adjustment and refines both joints and cameras, using per-view keypoint localization heatmaps as input.
Additional prior information, such as the fact that the triangle is equilateral, can be elegantly integrated in the meta-optimizer training.
}
\label{fig:method2d}
\end{figure*}

%% file: 3_method.tex
\newcommand{\joint}{\mathbf{j}}
\newcommand{\joints}{\mathbf{J}}
\newcommand{\real}{\mathbb{R}}
\newcommand{\camera}{\mathbf{c}}
\newcommand{\cameras}{\mathbf{C}}
\newcommand{\heatmap}{\mathbf{h}}
\newcommand{\heatmaps}{\mathbf{H}}
\newcommand{\keypoint}{\mathbf{k}}
\newcommand{\keypoints}{\mathbf{K}}
\newcommand{\image}{\mathcal{I}}
\newcommand{\identity}{\mathbf{I}}
\newcommand{\rotation}{\mathbf{R}}
\newcommand{\translation}{\mathbf{t}}
\newcommand{\scale}{s}
\newcommand{\iter}{{(i)}}
\newcommand{\iterZero}{{(0)}}
\newcommand{\iterNext}{{(i+1)}}
\newcommand{\updateNet}{\mathcal{F}_{\theta}}
\newcommand{\mixture}{\mathbf{g}}
\newcommand{\mixtures}{\mathbf{G}}
\newcommand{\roughjoint}{\mathbf{q}}
\newcommand{\roughjoints}{\mathbf{Q}}
\newcommand{\bonelength}{\mathbf{b}}
\newcommand{\bonelengths}{\mathbf{B}}

\section{Method}
\label{sec:method}
\noindent
As illustrated in \Figure{method2d}, given a collection of $\{\image_c\}$ images, we seek to optimize, up to a global rotation, scale, and shift:
\begin{itemize}[leftmargin=*]
\setlength\itemsep{-.3em}
\item $\joints=\{\joint_j \in \real^3\}_{j=1}^J$: the 3D coordinates of 3D body joints,
\item $\cameras=\{\camera_c \in \real^P\}_{c=1}^C$: the cameras parameters.
\end{itemize}
Having also observed:
\begin{itemize}[leftmargin=*]
\setlength\itemsep{-.3em}
\item $\heatmaps = \{\heatmap_c \in \real^{J \times H \times W} \}_{c=1}^C$: a set of 2D heatmaps of locations on images $\{\image_c\}$ captured using these cameras,
\end{itemize}
And assuming that, at training time, we are provided with:
\begin{itemize}[leftmargin=*]
\setlength\itemsep{-.3em}
\item $\keypoints = \{ \keypoint_{j,c}\}$: the ground truth 2D locations of the projection of joint $\joint_j$ in camera $\camera_c$.
\end{itemize}

\paragraph{Bayesian model}
Formally, assuming that heatmaps depend on camera parameters and joint positions $\joints$ only through 2D keypoint locations (i.e. $\Prob(\heatmaps|\keypoints, \joints, \cameras) = \Prob(\heatmaps|\keypoints)$), the joint distribution can be factorized as:
\begin{equation}
\Prob(\joints, \cameras, \keypoints, \heatmaps) = \Prob(\heatmaps|\keypoints) \, \Prob(\keypoints|\joints,\cameras) \, \Prob(\joints) \, \Prob(\cameras)
\label{eq:joint}
\end{equation}
Joints and keypoints are assumed to be related by:
\begin{equation}
\Prob(\keypoints|\joints,\cameras) = \prod_{j,c} \delta(\keypoint_{j,c} - \pi(\joint_j, \camera_c))
\label{eq:projection}
\end{equation}
where $\delta$ is the Dirac distribution, and $\pi(\joint, \camera)$ projects a joint $\joint$ to the 2D coordinates in camera $\camera$.
We use a weak-projection camera model
, hence, each camera is defined by a tuple of rotation matrix $\rotation$, pixel shift vector $\translation$, and single scale parameter $\scale$, \textit{i.e.} $\camera = [\rotation, \translation, \scale]$, and the projection operator is defined as $\pi(\joint, (\rotation, \translation, \scale)) = s \cdot \identity_{[0{:}1]} \cdot R \cdot \joint + t$ where $\identity_{[0{:}1]}$ is a truncated identity matrix that discards the third dimension of the multiplied vector.
This choice of the camera model simplifies initialization of camera parameters from single-view estimates (\Section{initialization}) and eliminates the ambiguity in focal length-vs-distance to the camera.
In Section~\ref{sec:results} we show experimentally what fraction of the final error comes from this camera model.

\paragraph{Inference task}
Our inference task is then to estimate the $\joints$ and $\cameras$ from observed heatmaps $\heatmaps$. 
We first introduce a probabilistic bundle adjustment formulation to handle joint position uncertainty, then propose a regression model that models complex interactions between joint positions and observed heatmaps.
The overall inference task can be framed as finding the maximum of the posterior probability of the pose and camera parameters given observed heatmaps, marginalized over possible keypoint locations:
\begin{equation}
\max_{\joints, \cameras}  \: \Prob(\joints, \cameras | \heatmaps) \!= \!\!\! \int \!\! \frac{\Prob(\keypoint | \heatmaps) \, \Prob(\keypoint|\joints,\cameras) \, \Prob(\joints) \, \Prob(\cameras)}{\Prob(\keypoint)} \diff \keypoint
\label{eq:main_objective}
\end{equation}
%
where, assuming that no prior information over camera parameters,  keypoint locations, and poses is given (i.e. constant~$\Prob(\cameras)$, $\Prob(\keypoints)$ and $\Prob(\joints)$) and using \eq{projection} we get:
\begin{equation}
\Prob(\joints, \cameras | \heatmaps) \propto \prod_{c,j} \Prob(\keypoint_{j,c}=\pi(\joint_j,\camera_c)|\heatmaps)
\label{eq:expansion}
\end{equation}
Further, assuming that each keypoint $\keypoint_{c,j}$ is affected only by a corresponding heatmap $\heatmap_{c,j}$,
and more specifically that the conditional probability density is proportional to the corresponding value of the heatmap:
\begin{equation}
\Prob(\keypoint_{j,c}|\heatmaps) = \Prob(\keypoint_{j,c}|\heatmap_{j,c}) \propto \heatmap_{j,c}[\keypoint_{j,c}]
\label{eq:naive_heatmap_keypoint}
\end{equation}
we get a probabilistic bundle adjustment problem:
\begin{equation}
\max_{\joints, \cameras}  \: \prod_{c,j} \heatmap_{j,c}[\pi(\joint_j,\camera_c)]
\label{eq:ba_objective}
\end{equation}
As we will show in Section~\ref{sec:results}, better estimation \textit{accuracy} with \textit{faster} inference time can be archived if assume 
that each keypoint can be affected by any heatmap via the following functional relation up to a normally distributed residual:
%
\begin{equation}
\Prob(\keypoints|\heatmaps, \theta) = \mathcal N( \keypoints \: | \: \pi(\joints_{\theta}(\heatmaps), \cameras_{\theta}(\heatmaps)), \identity)
\label{eq:parametric_objective}
\end{equation}
where $\joints_{\theta}, \cameras_{\theta}$ are joint and camera regression models (e.g. neural networks) parameterized by an unknown parameter $\theta$, and $\mathcal N$ is a multivariate normal density. Parameters of this model can be found via maximum likelihood estimation using observations from $\Prob(\keypoints, \heatmaps)$ available during training
\begin{align}
\theta_{\text{MLE}} &= \arg\max_{\theta} \:\: \Prob(\heatmaps, \keypoints|\theta) = \arg\max_{\theta}\Prob(\keypoints|\heatmaps, \theta) \\
&= \arg\min_{\theta} \:\: \mathbb{E}_{\keypoints, \heatmaps} \| \keypoints - \pi(\joints_{\theta}(\heatmaps), \cameras_{\theta}(\heatmaps))\|_2^2
\label{eq:theta_mle}
\end{align}
Then the test-time inference reduces to evaluation of the regression model at given heatmaps:
\begin{equation}
    \arg\max_{\joints, \cameras}  \: \Prob(\joints, \cameras | \heatmaps, \theta) = \joints_{\theta}(\heatmaps), \cameras_{\theta}(\heatmaps)
\end{equation}
Intuitively, the parametric objective enables complex interactions between all observed heatmaps and all predicted joint locations. The resulting model outperforms the probabilistic bundle adjustment both in terms of speed and accuracy, as we show in Section~\ref{sec:results}.

\paragraph{Solver}
To solve the highly non-convex problem in \eq{theta_mle}, and to do so \textit{efficiently}, we employ a modular \textit{two stages} approach; see~\Figure{method2d}:
\begin{description}[leftmargin=1em,font=\slshape]
\setlength\itemsep{-.3em}
\item[Stage 1 (S1): Initialization -- \Section{initialization}:] We first acquire an \textit{initial} guess $(\joints_\text{init}, \cameras_{\text{init}})$ using single-view 3D pose estimates for the camera configuration and the 3D pose by applying rigid alignment to per-view 3D pose estimates obtained using a pre-trained weakly-supervised single-view 3D network, e.g.~\cite{kocabas2019epipolar,wandt2020canonpose}
\item[Stage 2 (S2): Refinement -- \Section{refinement}:] We then train a neural network $f_{\theta}$ to predict a series of \textit{refinement} steps for camera and pose, staring from the initial guess so to optimize \eq{theta_mle}.
\end{description}
 
\paragraph{Advantages}
This approach has several key advantages:
\vspace{-.5em} 
\begin{enumerate}[leftmargin=*,label={\arabic*)}]
\setlength\itemsep{-.3em}
\item it \textit{primes} the refinement stage with a ``good enough'' guess to start from the correct basin of the highly non-convex pose likelihood objective given multi-view heatmaps;
\item it provides us with a \textit{modular} framework, letting us swap pre-trained modules for single-view 2D and 3D \textit{without} re-training the entire pipeline whenever a better approach becomes available;
\item the neural optimizer provides orders of magnitude \textit{faster inference} than classical iterative refinement, and allows the entire framework to be written within the same coherent computation framework~(i.e. neural networks vs. neural networks \textit{plus} classical optimization). 
\end{enumerate}

\subsection{Pre-processing}\label{subsec:preprocessing}
We assume that we have access to a 2D pose estimation model (e.g. PoseNet~\cite{papandreou2018personlab}) that produces 2D localization heatmaps $\heatmap_{j,c}$ for each joint $j$ from RGB image $\image_c$. 
We approximate each heatmap $\heatmap_{j,c}$ with an $M$-component mixture of spherical Gaussians $\mixture_{j,c}$.
This \textit{compressed} format 
reduces the dimensionality of the input to the neural optimizer (Section~\ref{sec:refinement}).
To fit parameters $\mixture_{j,c}$ of a mixture of spherical Gaussians to a localization 2D histogram $\heatmap_{j,c}$, we treat the heatmap as a regular grid of samples weighted by corresponding probabilities, and apply weighted EM algorithm \cite{frisch2021gaussian} directly to weighted samples, as described in the supplementary Section~\ref{subsec:em_algo}. 

\paragraph{Single-view pose estimation}
To initialize camera parameters via rigid alignment~(\Section{initialization}), we need a single-image 3D pose estimation model trained without 3D supervision~(e.g. EpipolarPose~\cite{kocabas2019epipolar}) that produces per-camera rough 3D pose estimates $\roughjoints = \{\roughjoint_{c,j}\}$ given an image~$\image_c$ from that camera.
These single-image estimates $\roughjoint_{c,j}$ are assumed to be in the camera frame, meaning that first two spatial coordinates of $\roughjoint_{c,j}$ correspond to \textit{pixel coordinates} of joint $j$ on image $\image_c$, and the third coordinate corresponds to its single-image relative zero-mean \textit{depth} estimate.

\input{figs/initialization}
\subsection{Initialization -- \Figure{initialization}}
\label{sec:initialization}
The goal of this stage is to acquire an initial guess for the 3D pose and cameras $(\joints_\text{init}, \cameras_{\text{init}})$ using single-view rough camera-frame 3D pose estimates $\roughjoints$ made by a model trained without 3D supervision \cite{kocabas2019epipolar,wandt2020canonpose}. We assume fixed initial parameters of the first camera
\begin{align}
\camera_0^{\text{init}} = (\rotation^\text{init}_0, \translation^\text{init}_0, \scale^\text{init}_0) = (\identity, \bar 0, 1)
\end{align}
and define initial estimates of rotations, scales and translations of remaining cameras as solutions the following orthogonal rigid alignment problem:
\begin{align}\label{eq:rigid_problem}
\argmin_{\rotation_c, \translation_c, \scale_c} \sum_{j} \|\roughjoint_{c,j} - (\scale_c \cdot \rotation_c \cdot \roughjoint_{0,j} + \identity_{[0:1]}^T \cdot \translation_c)\|^2
\end{align}
that can be solved using SVD of the outer product of mean-centered 3D poses~\cite{schonemann1966generalized}.
The initial guess for the 3D pose~$\joints_\text{init}$ then is the average of single-view 3D pose predictions~$\roughjoints$ rigidly aligned back into the first camera frame by corresponding estimated optimal rotations, scales and shifts:
\begin{align}
\joints^{\text{init}} = \tfrac{1}{C} \sum_c \rotation_c^T \cdot (\roughjoint_{c} - \identity_{[0:1]}^T \cdot \translation_c)) / s_c
\end{align}
%


\subsection{Refinement -- \Figure{refinement}}
\label{sec:refinement}
We train a neural network $f_{\theta}$ to predict a series of updates to 3D pose and camera estimates that leads to a refined estimate starting from the initialization from \Section{initialization}:
%
\begin{align}
\joints^\iterNext &= \joints^\iter + \diff \joints^\iter, &\joints^\iterZero = \joints_\text{init}
\\ 
\cameras^\iterNext &= \cameras^\iter + \diff \cameras^\iter, &\cameras^\iterZero = \cameras_\text{init}.
\end{align}
To ensure that inferred camera parameters $\cameras$ stay valid under any update $\diff \cameras$ predicted by a network, camera scale~(always positive) is represented in log-scale, and camera rotation uses a continuous 6D representation \cite{zhou2019continuity}.

At each refinement step $\diff \joints^\iter, \diff \cameras^\iter = \updateNet^\iter(\dots)$ the sub-network $\updateNet^\iter$ of the overall network $f_{\theta}$ is provided with as much information as possible to perform a meaningful update towards the optimal solution:
\begin{itemize}[leftmargin=*] \setlength\itemsep{-.3em}
\item $(\joints^\iter, \cameras^\iter)$ -- the current estimate to be refined;
\item $\mixtures {=} \{\mixture_{j,c}\}$ -- a collection of Gaussian mixtures compactly representing the heatmaps density distributions;
\item $\keypoints^\iter {=} \{ \keypoint_{j,c}^\iter = \pi(\joint_j^\iter, \camera_c^\iter)\}$ -- the set of projections of each joint $\joint{^\iter}$ into each camera frame $\camera^\iter$;
\item $\mathcal{L}(\joints^\iter, \cameras^\iter | \mixtures)$ -- the likelihood of the current estimate of joints given the heatmap mixture parameters.
\end{itemize}
These learnt updates seek to minimize the L2 distance between predicted and ground truth 2D coordinates of keypoints in each frame, mirroring the maximum likelihood objective \eq{theta_mle} we defined earlier:
\begin{equation}
\arg\min_\theta \: \mathcal{L}_\keypoint(\theta) = \sum_\iter \sum_{j,c}
\| \keypoint_{j,c}^\iterNext - \keypoint_{j,c}^\text{gt} \|_2^2
\label{eq:loss_reprojection}
\end{equation}
where, in practice, we train refinement steps~$\mathcal \updateNet^\iter$ progressively, one after the other.

\input{figs/architecture}
\paragraph{Architecture design}
The architecture of $\updateNet$ needs to be very carefully designed to respect the symmetries of the problem at hand.
The inferred updates to $\joints^\iterNext$ ought to be \textit{invariant} to the order of cameras, while updates to $\cameras^\iterNext$ ought to be \textit{permutation-equivariant} w.r.t. the current estimates of~$\cameras^\iter$, rows of~$\keypoints^\iter$, and Gaussian mixtures~$\mixtures$. 
Formally, for any inputs and permutation of cameras $\sigma$:
\begin{align}
\diff \joints, \diff \cameras  = \mathcal F_{\theta}(\joints^\iter, \cameras^\iter, \mixtures, \keypoints^\iter, \mathcal L) \\
\diff \joints', \diff \cameras'  = \mathcal F_{\theta}(\joints^\iter, \cameras^\iter_\sigma, \mixtures_\sigma, \keypoints^\iter_\sigma, \mathcal L) 
\end{align}
we need to guarantee that $\diff \joints = \diff \joints'$ and $\diff \cameras = \diff \cameras'_\sigma$.
To archive this, we concatenate view-invariant inputs $\joints^\iter$ and $\mathcal L$ to each row of view-dependant inputs $\cameras^\iter, \mixtures, \keypoints^\iter$, pass them though a permutation-equivariant MLP~\cite{deng2021vector,joseph2019momen} with aggregation layers concatenating first and second moments of feature vectors back to these feature vectors, and apply mean aggregation and a non-permutation-equivariant MLP to get the final pose update, as illustrated in~\Figure{architecture}.

\paragraph{Limitations}
We assume a weak camera model, making our method less accurate on captures shot using wide-angle (short-focus) lenses.
To achieve best performance, our method requires accurate 2D keypoint ground truth for training, but we also report performance without using GT keypoints during training (\Table{h36m_ablation}).
We implicitly assume that the subject is of comparable size (in pixels) across all views, and expect re-weighting of different components of the reprojection loss \eq{loss_reprojection} might otherwise be necessary. While tracking and pose estimation can be used for malicious purposes, such as mass surveillance, we believe that societal benefits brought by such technological advances~\cite{saraee2019exercisecheck,gu2019home,feng2021hgaze} outweigh possible abuse.

\subsubsection{Pose prior (i.e.~``bone-length'' experiment)}
\label{sec:poseprior}
We illustrate the modularity of our solution by effortlessly injecting a \textit{subject-specific} bone-legth prior into our meta-optimizer. Given two joints $\joint_n$ and $\joint_m$ connected in the human skeleton $\mathcal{E}$ by an edge $e=(n,m)$, we define the bone length $b_e(\joints)=\|\joint_n - \joint_m\|_2$. However, as our bundle adjustment is performed \textit{up to scale} we ought to define \textit{scale-invariant} bone lengths ${b}^N(\joints) = b(\joints) / \hat \mu(b(\joints))$ by expressing length of each bone relative to the average length of other bones~$\hat \mu(b)=(\sum_e b_e)/|\mathcal E|$. If we assume that during training and inference we observe noisy normalized bone-lenghs vectors $\bonelengths = b^N(\joints) + \varepsilon$, where~$\varepsilon \sim \mathcal N(0, \sigma_b^2 \identity)$.
Then, the joint probability \eq{joint} becomes:
\begin{equation*}
    \Prob(\joints, \cameras, \keypoints, \heatmaps, \bonelengths) = \Prob(\bonelengths | \joints) \, \Prob(\heatmaps|\keypoints) \, \Prob(\keypoints|\joints,\cameras) \, \Prob(\joints) \, \Prob(\cameras)
\end{equation*}
and our parametric likelihood \eq{parametric_objective} becomes:
\begin{equation*}
    \Prob(\keypoints|\heatmaps, \bonelengths, \theta) \propto 
    \Prob(\keypoints|\heatmaps, \theta) \, \cdot \, \mathcal N(b^N(\joints_{\theta}(\heatmaps, \bonelengths))| \bonelengths, \sigma_b^2 \identity)
\end{equation*}
and its parameters $\theta$ can be estimated equivalently to \eq{theta_mle} via maximum over $\Prob(\keypoints, \heatmaps, \bonelengths | \, \theta)$ using observations from $\Prob(\keypoints, \heatmaps, \bonelengths)$ available during training, effectively resulting in an additional loss term penalizing derivations of bone lengths of predicted poses from provided bone lengths:
\begin{equation}
\mathcal{L}_b(\theta) = \sum_\iter \left\| b^N(\joints^\iterNext) - \bonelengths \right\|^2_2.
\end{equation}



%% file: figs/initialization.tex
\begin{figure*}
\begin{minipage}{\columnwidth}
\begin{center}
\includegraphics[width=\linewidth,trim=0 2.9in 5.5in 0,clip]{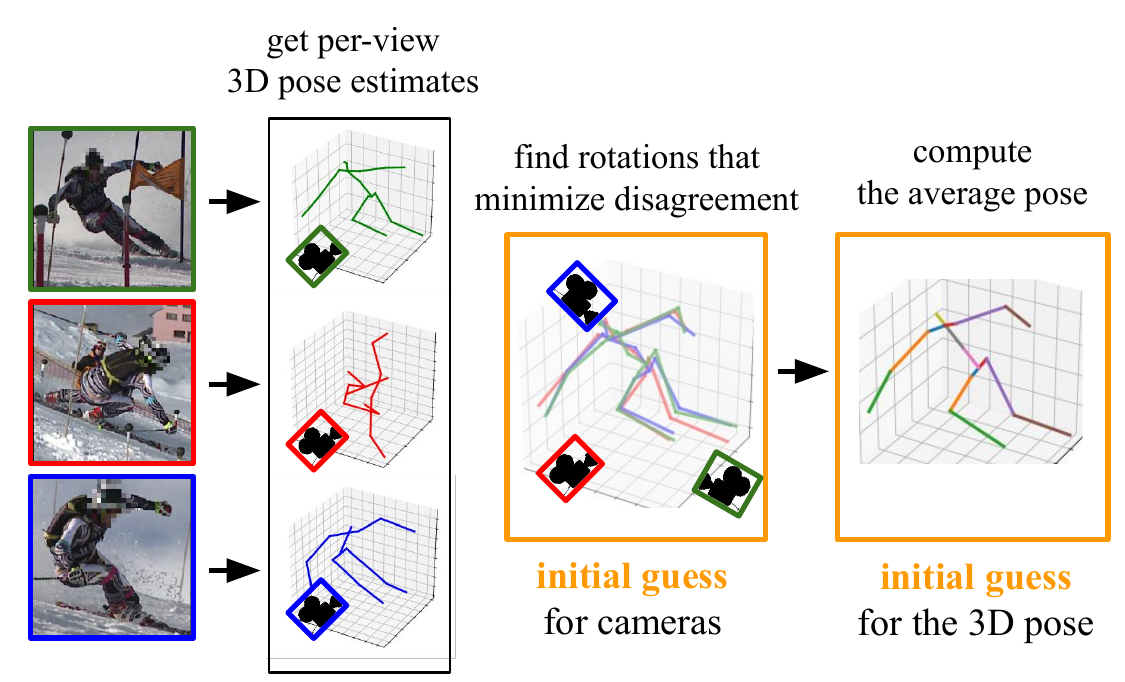}
\end{center}
\vspace{-10px}
\captionof{figure}{
\textbf{Initialization} --
We form an initial guess for the 3D pose \textit{and} the cameras by taking the mean of rigid aligned 3D poses estimated from each RGB image using an external single-view weakly-supervised 3D pose estimation network \cite{kocabas2019epipolar,wandt2020canonpose}.}
\label{fig:initialization}
\end{minipage}
\hfill
\begin{minipage}{\columnwidth}
\begin{center}
\includegraphics[width=\linewidth,trim=0 2.9in 5.4in 0,clip]{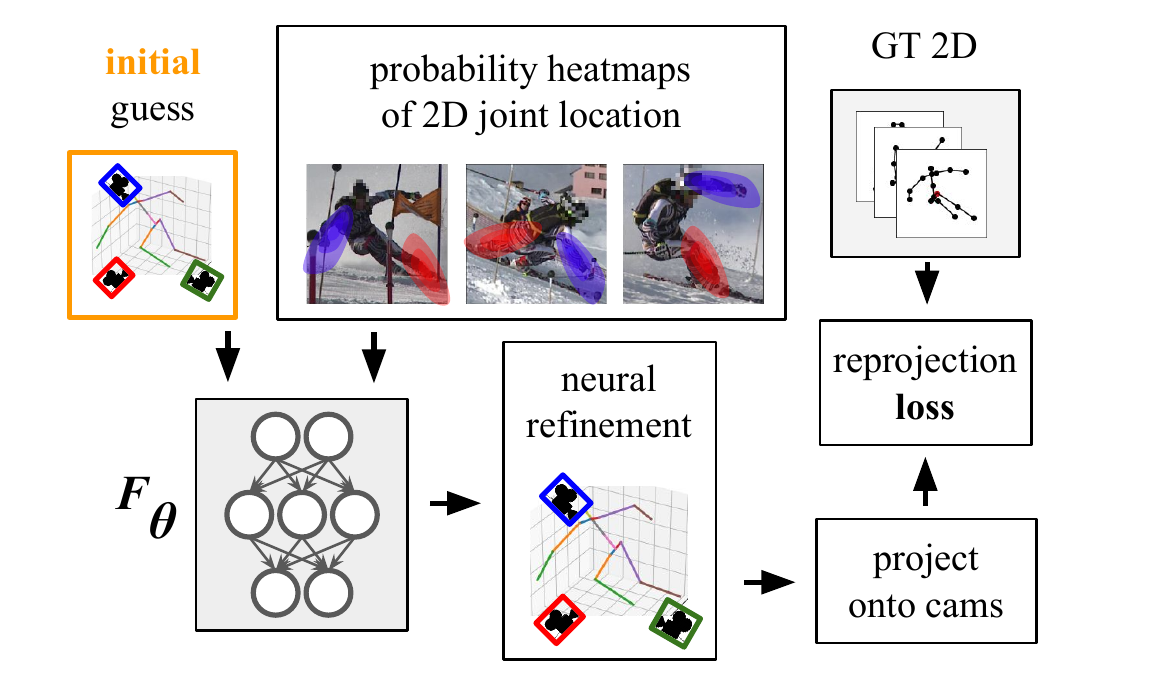}
\end{center}
\vspace{-10px}
\captionof{figure}{
\textbf{Refinement} --
We train a \textit{neural optimizer} $f_{\theta}$ to predict iterative refinement that minimizes the reprojection error with the ground truth re-projection, using the current guess and joint heatmaps as an input.
During inference, we \textit{do not} need ground truth 2D projections.
}
\label{fig:refinement}
\end{minipage}
\end{figure*}

%% file: figs/architecture.tex
\begin{figure*}
\centering
\includegraphics[width=2\columnwidth,trim=0 4in 1in 0,clip]{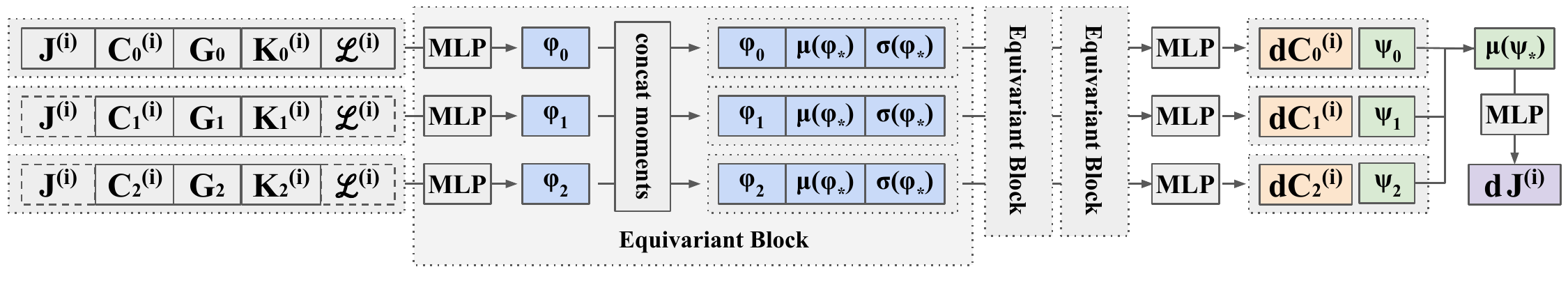}
\vspace{-10px}
\caption{
\textbf{Architecture} --
In order for predicted updates to respect symmetries of the problem at hand, we copy and concatenate view-invariant inputs (current pose estimate, average heatmap likelihood - dashed line) to each row of view-specific inputs (current cameras and joint projections, heatmaps), pass them though a Permutation-Equivariant MLP Block shown above.
To get permutation-invariant final pose update we additionally apply MLP to averaged output pose embeddings.
}
\label{fig:architecture}
\end{figure*}

%% file: 4_experiments.tex
\input{figs/qualitative}
\vspace{-3px}\section{Experiments}

In this section, we specify datasets and metrics we used to validate the performance of the proposed method and a set of baselines and ablation experiments we conducted to evaluate the improvement in error provided by each stage and each supervision signal.

\paragraph{Data} 
We evaluated our method on Human3.6M~\cite{ionescu2013human36m} dataset with four fixed cameras and a more challenging SkiPose-PTZ~\cite{rhodin2018skipose} dataset with six \textit{moving} pan-tilt-zoom cameras.
We used standard train-test evaluated protocol for H36M \cite{iskakov2019learnable,kocabas2019epipolar} with subjects 1, 5, 6, 7, and 8 used for training, and 9 and 11 used for testing. We additionally pruned the H36M dataset by taking each 16-th frame from it, resulting in 24443 train and 8516 test examples, each example containing information from four cameras.
We evaluated our method on the subset (1035 train / 230 test) of SkiPose~\cite{rhodin2018skipose} that was used in CanonPose~\cite{wandt2020canonpose} that excludes 280 examples with visibility obstructed by snow. 
In each dataset, we used the first 64 examples from the train split as a validation set.

\paragraph{Metrics}
We report Procrustes aligned Mean Per Joint Position Error (PMPJPE) and Normalized Mean Per Joint Position Error (NMPJPE) that measure the L2-error of 3D joint estimates after applying the optimal rigid alignment (including scale) to the predicted 3D pose and the ground truth 3D pose (for NMPJPE), or only optimal shift and scale (for PMPJPE). 
We also report the total amount of time ($\Delta t$) it takes to perform 3D pose inference from multi-view RGB.

\paragraph{Baselines} 
On H36M we lower-bound the error 
with the state-of-the-art fully-supervised baseline of~\citet{iskakov2019learnable} that uses \textit{ground truth} camera parameters to aggregate multi-view predictions during inference.
We also compare the performance of our method to methods that use multi-view 2D supervision during training but only perform inference on a single view at a time: self-supervised EpipolarPose~(EP) \cite{kocabas2019epipolar} and CanonPose~(CP) \cite{wandt2020canonpose}, as well as the weakly supervised baselines of \citet{iqbal2020weakly} and \citet{rhodin2018skipose}. 
On SkiPose we compared our model with the only two baselines available in the literature: CanonPose~\cite{wandt2020canonpose} and~\citet{rhodin2018skipose}.
We did not evaluate EpipolarPose on SkiPose because it requires fixed cameras to perform the initial self-supervised pseudo-labeling.
We did not evaluate~\citet{iqbal2020weakly} on SkiPose because no code has been released to date and authors did not respond to a request to share code.

We also compared our method against the ``classical'' bundle adjustment initialized with ground truth extrinsic camera parameters of all cameras, and set fixed GT intrinsics, therefore putting it into \textit{unrealistically favorable} conditions.
We used the well-tested implementation of bundle adjustment in AniPose~\cite{karashchuk2020anipose} that uses an adapted version of the 3D registration algorithm of~\citet{zhou2016fast}.
This approach takes point estimates of keypoint locations as an input (i.e. no uncertainty) and iteratively detects outliers and refines camera parameters and joint 3D positions using the second-order Trust Region Reflective algorithm \cite{byrd1988approximate,branch1999subspace}. 

\paragraph{Architecture}
For monocular 2D pose estimation, we used the stacked hourglass network \cite{newell2016stacked} pre-trained on COCO pose  dataset \cite{guler2018densepose}. 
For monocular 3D estimation in Stage 1, we applied EpipolarPose \cite{kocabas2019epipolar} on Human3.6M and CanonPose \cite{wandt2020canonpose} on SkiPosePTZ.
We note that differences in the joint labeling schemes used by these monocular 3D methods and our evaluation set do not affect the quality of \textit{camera} initialization we acquire via rigid alignment, as long as monocular 3D estimates for all views follow a consistent labeling scheme. 
Each neural optimizer step is trained separately, and stop gradient is applied to all inputs.
We refer our readers to Section~\ref{subsec:ref_perform} in supplementary for a more detailed description of all components we used to train our neural optimizer and their reference performance. 

%% file: figs/qualitative.tex
\begin{figure*}
\centering
\arrayrulecolor{gray}%
\setlength{\arrayrulewidth}{1pt}%
\vspace{-10px}
\includegraphics[width=\linewidth,trim=0 2.5in 0.1in 0,clip]{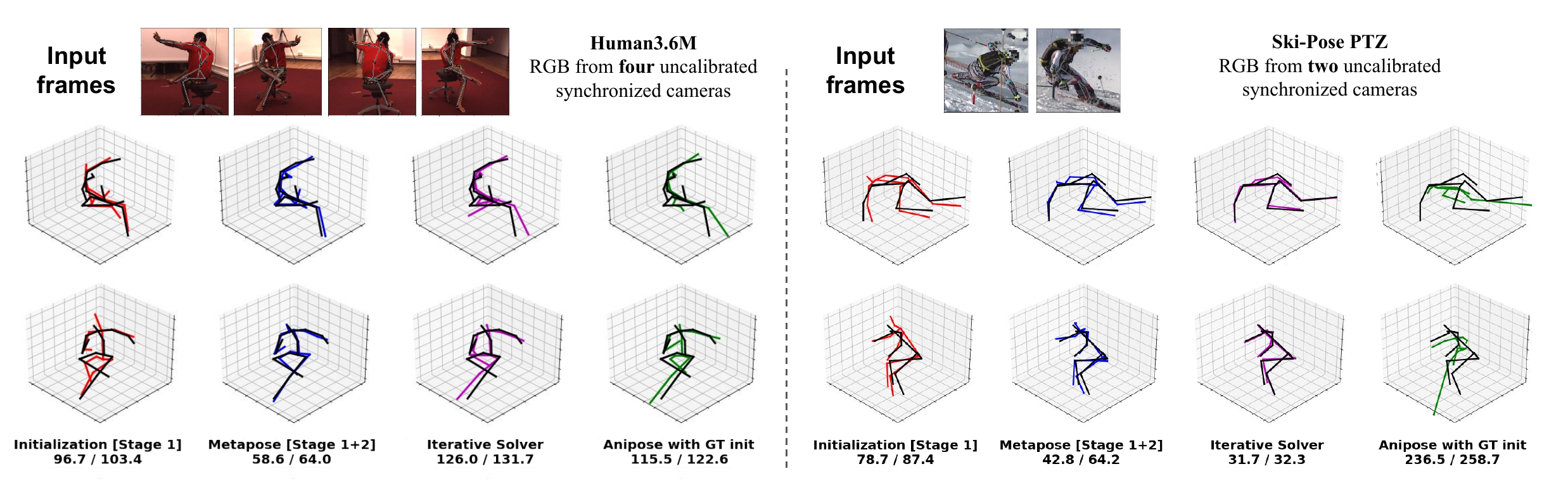}
\vspace{-8px}
\caption{
\textbf{Qualitative Results} --
The top row shows input frames we used for pose estimation, overlayed with the GT pose (\textbf{black}). Two bottom rows show predictions made by evaluated methods on \textbf{H36M with four cameras} (left) and \textbf{SkiPose with two cameras} (right). We include predictions for {\color{red}Initialization} (Stage 1), {\color{blue}MetaPose} (Stage 1+2), MetaPose with an {\color{Plum}Iterative Refinement} (S1+IR), and {\color{OliveGreen}AniPose} initialized with GT. We also provide errors in the format: PMPJPE/NMPJPE.
A video demonstration of qualitative results across both datasets can be found in the supplementary material or at this link: \href{https://bit.ly/cvpr22_6639}{https://bit.ly/cvpr22\_6639}.
}\vspace{-3px}
\label{fig:qualtitative}
\end{figure*}

%% file: 5_results.tex
\input{tables/results_h36m_ski}
\section{Results -- \Table{quantitative}}\label{sec:results}
\input{tables/results_ablations_new}

The proposed method (MetaPose S1+S2)  outperforms the classical bundle-adjustment baseline initialized with ground truth cameras (AniPose \cite{karashchuk2020anipose} w/ GT) by +40mm on H36M with four cameras, and +8mm on SkiPose with six cameras. With fewer cameras the performance gap increases further.
MetaPose also outperforms semi-, weakly-, and self-supervised baselines reported in prior work~\cite{iqbal2020weakly,rhodin2018skipose,wandt2020canonpose,kocabas2019epipolar} by more then 10mm. 
We would like to re-iterate core advantages of the proposed method beyond its high performance, namely:
\CIRCLE{1} that Stage 1 \textit{primes} the neural optimizer with a good enough initialization that leads it to a good solution;
\CIRCLE{2} that our solution is \textit{modular} enabling swapping existing priming and pose estimation networks, as well as additional losses, and re-training only the neural optimizer;
\CIRCLE{3} that our method achieves \textbf{lower latency} then both classical and~(GPU-accelerated) probabilistic bundle adjustment. We expand upon these and other related finding in the next subsection.

\subsection{Ablations}\label{subsec:ablation}

\paragraph{Iterative refiner}
We measured the speed gain we get from using the neural optimizer $f_{\theta}$ by replacing Stage 2 with a test-time GPU-accelerated gradient descent~(Adam~\cite{kingma2014adam}) over the probabilistic bundle adjustment objective \eq{ba_objective} with GMM-parameterized heatmaps.
Section 1 in \Table{h36m_ablation} shows that the proposed method (S1+S2) is up to \textit{seven times} faster than the iterative refinement (S1+IR), and is at least 10mm more accurate.
We also measured the contribution of keypoint supervision towards prediction accuracy of S2  compared to iterative refinement.
To do that, we trained Stage 2 to minimize the same GMM-parameterized probabilistic bundle adjustment objective \eq{ba_objective} instead of the re-projection loss \eq{loss_reprojection}.
The resulting self-supervised model (S1+S2/SS) outperforms the iterative refinement, suggesting that the proposed architecture regularizes the pose estimation problem.
Note that our self-supervised results also outperform prior work that uses weak- and self-supervision \cite{iqbal2020weakly,kocabas2019epipolar,wandt2020canonpose}.

\paragraph{Random initialization} We measured the effect of replacing single-view pose estimates $\roughjoint_{c,j}$ used to initialize the pose and cameras in Stage~1 with random Gaussian noise.
Section~2 in \Table{h36m_ablation} shows that while the neural optimizer (RND+S2) is more resilient to poor initialization then the classical one (RND+IR), a good initialization is necessary to achieve the state-of-art performance (S1+S2). Moreover, marginally better results with GT initialization (GT+IR) show that the proposed initialization already brings the optimizer in the neighbourhood of the correct solution, and that further improvement in the quality of the initial guess will not provide significant gains in accuracy.

\paragraph{Non-equivariant network}
We measured the effect of letting the model ``memorize'' the camera order by replacing equivariant blocks with MLPs that receive multi-view information as a single concatenated vector.
The resulting model (S2/MLP) achieved marginally better performance on H36M and marginally worse performance on SkiPose (Table~\ref{tab:skipose_mlp}), likely due to fixed cameras positions in H36M and moving cameras in SkiPose.

\paragraph{Bone lengths} We trained a model with an additional bone length prior (Sec.~\ref{sec:poseprior}) that improved PMPJPE with two cameras by 7mm. The two camera setup is ill-conditioned, hence can better exploit priors like bone lengths.

\paragraph{Inputs of neural optimizer}
Unsurprisingly, among all inputs passed to the neural optimizer, heatmap $\heatmaps$ contributed most to the final performance, but all components were necessary to achieve the state-of-art performance; see \Table{neural_abl} in supplementary.

\paragraph{Further ablations (supplementary)}
The teacher-student loss proposed by~\citet{ma2020deepoptimizer} to draw predicted solutions into the basin of the right solution \textit{hurts} the performance in all experiments (\Table{teacher_student}), suggesting that Stage 1 already provides good-enough initialization to start in the correct basin of the objective.
We also ran the iterative refiner from ground truth initialization with re-projection losses with different camera models: results suggests that the weak camera model contributed to 10-15mm of error on H36M and no error on SkiPose; see~\Table{camera_model}. 
The performance of MetaPose on H36M starts to severely deteriorate at around 5\% of the training data; see~\Table{sup_h36m_size}.
Replacing GMM with a single Gaussian decreased the performance only in two-camera H36M setup by 4mm, and did not significantly influence the performance in other cases; see~\Table{gmm}.

%% file: tables/results_h36m_ski.tex
\begin{table}[t]
\begin{center}
\inctabcolsep{-1pt} {
\begin{tabular}{lccccc}\toprule
\multicolumn{1}{l}{\multirow{2}{*}{Method}} & 
\multicolumn{2}{c}{PMPJPE$\downarrow$} & \multicolumn{2}{c}{NMPJPE$\downarrow$} &  \multirow{2}{*}{\stackanchor{$\Delta t$}{[s]}} \\
\cmidrule(lr){2-3} \cmidrule(lr){4-5}
& 4 & 2 & 4 & 2 \\
\midrule
\rowcolor[gray]{0.9}
Isakov et al. \cite{iskakov2019learnable} & 20 & - & - & - & - \\ 
\midrule
AniPose \cite{karashchuk2020anipose} w/ GT & 75 & 167 & 103 & 230 & 7.0 \\
Rhodin et al. \cite{rhodin2018skipose} & 65 & - & 80 & - & - \\
CanonPose \cite{wandt2020canonpose} & 53 & - & 82 & - & - \\
EpipolarPose (EP) \cite{kocabas2019epipolar} & 71 & - & 78 & - & - \\
Iqbal et al. \cite{iqbal2020weakly} & 55 & - & 66 & - & - \\
\midrule
\textbf{MetaPose} (S1) & 74 & 87 & 83 & 95 & \textbf{0.2} \\
\textbf{MetaPose} (S1+S2) & \textbf{32} & \textbf{44} & \textbf{49} & \textbf{55} & 0.3 \\
\bottomrule
\end{tabular}

\bigskip
\begin{tabular}{lccccc}\toprule
\multicolumn{1}{l}{\multirow{2}{*}{Method}} & 
\multicolumn{2}{c}{PMPJPE$\downarrow$} & \multicolumn{2}{c}{NMPJPE$\downarrow$} &  \multirow{2}{*}{\stackanchor{$\Delta t$}{[s]}} \\
\cmidrule(lr){2-3} \cmidrule(lr){4-5}
& 6 & 2 & 6 & 2 \\
\midrule
AniPose \cite{karashchuk2020anipose} w/ GT & 50 & 62 & 221 & 273 & 7.0 \\
Rhodin et al. \cite{rhodin2018skipose} & - & - & 85 & - & - \\
CanonPose (CP) \cite{wandt2020canonpose} $\ $ & 90 & - & 128 & - & - \\
\midrule
\textbf{MetaPose} (S1) & 81 & 86 & 140 & 144 & \textbf{0.3} \\
\textbf{MetaPose} (S1+S2) & \textbf{42} & \textbf{50} & \textbf{53} & \textbf{59} & 0.4 \\
\bottomrule
\end{tabular}
}
\end{center}
\vspace{-1em}
\caption{
\textbf{Quantitative comparison to prior work} --
Performance of different methods with four and two cameras on \textbf{Human3.6M} (top) and six and two cameras \textbf{SkiPose-PTZ} (bottom), Procrustes and Normalized MPJPE in millimeters, inference time in seconds. See Tables \ref{tab:latency} and \ref{tab:quantitative_extended} in the supplementary material for an extended comparison and breakdown of runtime performance.\vspace{-10px}
}
\label{tab:quantitative}
\end{table}

%% file: tables/results_ablations_new.tex
\begin{table}[t]
\centering
\inctabcolsep{-1pt} {
\begin{tabular}{lccccc}\toprule
\multicolumn{1}{l}{\multirow{2}{*}{Method}} & 
\multicolumn{2}{c}{PMPJPE$\downarrow$} & \multicolumn{2}{c}{NMPJPE$\downarrow$} &  \multirow{2}{*}{\stackanchor{$\Delta t$}{[s]}} \\
\cmidrule(lr){2-3} \cmidrule(lr){4-5}
& 4 & 2 & 4 & 2 \\
\midrule
\textbf{MetaPose} (S1+S2) & \textbf{32} & \textbf{44} & \textbf{49} & \textbf{55} & 0.3 \\
\textbf{MetaPose} (S1+IR) & 43 & 53 & 66 & 75 & {2.0} \\
\textbf{MetaPose} (S1+S2/SS) & 39 & 50 & 56 & 63 & 0.3 \\
\midrule
\textbf{MetaPose} (S1+S2) & \textbf{32} & \textbf{44} & \textbf{49} & \textbf{55} & 0.3 \\
\textbf{MetaPose} (RND+S2) & 36 & 51 & 52 & 64 & 0.3 \\
\textbf{MetaPose} (S1+IR) & 43 & 53 & 66 & 75 & 2.0 \\
\textbf{MetaPose} (RND+IR) & 200 & 385 & 265 & 444 & 2.0 \\
\textbf{MetaPose} (GT+IR) & 40 & 48 & 63 & 68 & 2.0 \\
\midrule
\textbf{MetaPose} (S1+S2) & 32 & \textbf{44} & 49 & \textbf{55} & 0.3 \\
\textbf{MetaPose} (S1+S2/MLP) & \textbf{30} & \textbf{44} & \textbf{47} & 58 & 0.3 \\
\midrule
\textbf{MetaPose} (S1+S2) & 32 & 44 & \textbf{49} & 55 & 0.3 \\
\textbf{MetaPose} (S1+S2/BL) & \textbf{30} & \textbf{37} & 50 & \textbf{54} & 0.3 \\

\bottomrule
\end{tabular}
}
\caption{\textbf{Ablations on H36M}. Notation consistent with \Table{quantitative}. \label{tab:h36m_ablation}}
\end{table}

%% file: 6_conclusions.tex
\section{Conclusions}


In this paper, we propose a new modular approach to 3D pose estimation that requires only 2D supervision for training and significantly improves upon the state-of-the-art by fusing per-view outputs of singe-view modules with a simple view-equivariant neural network. Our modular approach not only enables practitioners to analyze and improve the performance of each component in \textit{isolation}, and channel future improvements in respective sub-tasks into improved 3D pose estimation ``for free'', but also provides a common ``bridge'' that enables easy inter-operation of different schools of thought in 3D pose estimation -- enriching both the ``end-to-end neural world'' with better model-based priors and improved interpretability, and the ``iterative refinement world'' with better-conditioned optimization problems, transfer-learning, and faster inference times. We provide a detailed ablation study dissecting different sources of the remaining error, suggesting that future progress in this task might come from the adoption of a full camera model, further improvements in 2D pose localization, better pose priors and incorporating temporal signals from video data. 

%% file: 8_ack.tex
\section{Acknowledgement}
We would like to thank Bastian Wandt, Nori Kanazawa and Diego Ruspini for help with CanonPose \cite{wandt2020canonpose}, stacked hourglass pose estimator, and interfacing with AniPose, respectively.

%% file: 7_supplementary.tex
\input{tables/results_neural_abl}
\input{tables/sup_latency_breakdown}
\input{tables/sup_mlp_ski_h36m}
\input{tables/sup_results_h36m_ski_extended}
\input{tables/sup_bone_lengths}
\input{tables/sup_teacher_student}
\input{tables/sup_monocular_details}
\input{tables/sup_camera_model}
\input{tables/sup_h36m_size}
\setcounter{table}{12}
\input{tables/sup_gmm}

\section{Supplementary}
\subsection{Ablation Tables}
\begin{itemize}[nosep,leftmargin=*]
    \item \textbf{Table \ref{tab:neural_abl}} shows the performance of the neural optimizer trained with different subsets of inputs;
    \item \textbf{Table~\ref{tab:latency}} shows the latency breakdown across model components and models;
    \item \textbf{Table~\ref{tab:skipose_mlp}} shows that the performance of equivariant (S1+S2) and non-equivariant (S1+S2/MLP) models differs by at most 4mm on both datasets;
    \item \textbf{Table~\ref{tab:quantitative_extended}} shows that MetaPose outperforms prior work with corresponding supervision signals;
    \item \textbf{Table~\ref{tab:bone_lengths}} shows that the personalized bone length prior improves the performance of both the iterative and neural refiners in the majority of cases;
    \item \textbf{Table~\ref{tab:teacher_student}} shows that the student-teacher loss inspired by \citet{ma2020deepoptimizer} to draw the predicted solution into the correct basin of the loss hurts the performance in all cases;
    \item \textbf{Table~\ref{tab:mono_data}} summarizes reference performance of monocular pose estimation components across different splits of data (train, val, test) for reproducibility, and shows strong overfitting on SkiPose;
    \item \textbf{Table~\ref{tab:camera_model}} shows that at least 20mm of error is due to imperfect heatmaps, up to 10mm is due to the weak camera model, and only up to 3mm is due to imperfect init;
    \item \textbf{Table~\ref{tab:sup_h36m_size}} shows that on H36M with just 1/5th of the entire training dataset (i.e. 5k labeled training samples, each sample containing several cameras) we can get a model that achieves PMPJPE within 5-10mm of the performance we achieve on full data.
    \item \textbf{Table~\ref{tab:gmm} }shows the effect of varying the number of Gaussian mixture components on the performance of different methods.
\end{itemize}

\subsection{Weighted EM-algorithm for spherical GMM}\label{subsec:em_algo}
We used grid points $x_i$ weighted by corresponding probabilities $p_i$ to fit a GMM to a 2D probability heatmap. Following \citet{frisch2021gaussian} on each step $t=0\dots T$ of the EM algorithm we performed usual (non-weighted) E-step to compute the new assignment matrix $\eta_{i,m}^{(t+1)}$ between points $x_i$ and spherical clusters $m=0\dots M$ with means $\mu_m^{(t)}$, and standard variations $\sigma_m^{(t)}$, and weights $w_m^{(t)}$, followed by a \textbf{weighted} M-step:
\begin{gather*}
    w^{(t+1)}_m = \frac{\sum_i \eta^{(t+1)}_{i,m} p_i}{\sum_{m'} \sum_i \eta^{(t+1)}_{i,{m'}} p_i} \\
    \mu^{(t+1)}_m = \frac{\sum_i \eta^{(t+1)}_{i,m} p_i x_i}{\sum_i \eta^{(t+1)}_{i,m} p_i} \\
    \sigma^{(t+1)}_m = \sqrt{\frac{\sum_i \eta^{(t+1)}_{i,m} p_i ||x_i - \mu^{(t+1)}_m||_2^2}{\sum_i \eta^{(t+1)}_{i,m} p_i}}
\end{gather*}
\subsection{Implementation Details}\label{subsec:ref_perform}

\paragraph{Architecture}  For monocular 2D pose estimation we used the stacked hourglass network \cite{newell2016stacked} pre-trained on COCO pose 
dataset \cite{guler2018densepose}. We additionally trained a linear regression adapter to convert between COCO and H36M label formats (see supplementary Figure~\ref{fig:fig_label_formats} for labeling format comparison). The resulting procedure yields good generalization on H36M, as shown in supplementary Table~\ref{tab:mono_data}). The COCO-pretrained network generalized very poorly to SkiPosePTZ dataset because of the visual domain shift, so we fine-tuned the stacked hourglass network using ground truth 2D labels.
For monocular 3D estimates used in Stage 1, we applied EpipolarPose \cite{kocabas2019epipolar} on Human3.6M and CanonPose \cite{wandt2020canonpose} on SkiPosePTZ. 
We would like to note that, despite the significant shift in the labeling format between predictions of these monocular 3D methods and the format used in datasets we used for evaluation, this does not affect the quality of camera initialization we acquired via rigid alignment. Similar to prior work \cite{ma2020deepoptimizer}, each ``neural optimizer step'' is trained separately, and the fresh new neural net is used at each stage, and stop gradient is applied to all inputs. %
For MLP architecture, we used $L$ fully-connected 512-dimensional layers followed by a fully-connected 128-dimensional, all with selu with L=4 for H36M and L=2 for SkiPose. For equivalent network, the optimal network for H36M had following layers: [512, 512, CC, 512, 512, CC, 512] and for SkiPose had following layers: [512, 512, CC, 512, 512, CC, 512, 512, CC, 512, 512, CC, 512] - where CC corresponds to concatenation of first two moments and numbers correspond to dense layers with selu.
We re-trained each stage multiple times until the \textit{validation} PMPJPE improved or the total number of ``stage training attempts'' exceeded 100. 

\paragraph{Hyperparameters} 
We used Adam \cite{kingma2014adam} optimizer with learning rate \texttt{1e-2} for 100 steps for exact refinement, and \texttt{1e-4} for the neural optimizer. 

\paragraph{Reference 2D performance}\label{sec:extended_train_test_scores}
Tables~\ref{tab:mono_data} shows performance of 2D pose prediction networks and the resulting MetaPose network on different splits of different datasets. It shows that both the 2D network and MetaPose to certain degree overfit to SkiPose because of its smaller size. 

\input{figs/fign_labl_fmt}

Videos with test predictions can be found in the attached video file and following this link: \href{https://bit.ly/cvpr22_6639}{https://bit.ly/cvpr22\_6639}.


\subsection{Closed Form Expressions for Stage 1}\label{subsec:closed_form_stage1}
Below we describe the solution to the rigid alignment problem (\ref{eq:rigid_problem}) for monocular 3D pose estimates $\roughjoint_{c}$ and inferred weak camera parameters from them. Assume that we have monocular 3D predictions $\roughjoint_{c}$ in frame of the camera $c$. The parameters of the first camera are assumed to be known and fixed
$$R^{(0)}_{\text{init}}{=}I, \bm{t}^{(0)}{=}\bar 0, s^{(0)}{=}1$$
whereas the rotation of other cameras are inferred using optimal rigid alingment $R^{(c)}_{\text{init}} = (U^{(c)})^T V^{(c)}$
where
$$U^{(c)}, \Lambda, V^{(c)} = \operatorname{SVD}(\operatorname{centered}(\roughjoint_{c}) \cdot \operatorname{centered}(\roughjoint_{0}))^T)$$
The scale $s$ and shift $\bm{t}$ can be acquired by comparing the original monocular $\roughjoint_{c, [:,0:2]}$ in pixels to $[R^{(c)}_{\text{init}} \operatorname{centered}(\roughjoint_{0})]_{[:,0:2]}$ rotated back into each camera frame, for example:
\begin{align}
s^{(c)}_{\text{init}} = \frac{|| [(R^{(c)}_{\text{init}})^T \operatorname{centered}(\roughjoint_{0})]_{[:,0:2]} ||}{|| \operatorname{centered}(\roughjoint_{c})_{[:,0:2]} ||}
\end{align}
\begin{align}
\bm{t}^{(c)}_{\text{init}} = & \hat \mu([(R^{(c)}_{\text{init}})^T \operatorname{centered}(\roughjoint_{0})]_{[:,0:2]}) - \hat \mu([\roughjoint_{c}]_{[:,0:2]}))
\end{align}
where $\hat \mu(\bm{a}) = (\sum_k^K \bm{a}_{k}) / K$ is the center of the 3D pose and $\operatorname{centered}(\bm{a})_{k} = (\bm{a}_k - \hat \mu(\bm{a}))$ and the initial pose estimate is the average of aligned, rotated and predictions from other cameras. The initial guess for the pose is the average of all monocular poses rotated into the first camera frame:
\begin{equation}
\bm{J}_{\text{init}} = \frac{1}{C}\sum_{c=0}^C (s^{(c)}_{\text{init}} \cdot R^{(c)}_{\text{init}} \operatorname{centered}(\roughjoint_{c})) + \hat \mu(\roughjoint_{0})
\end{equation}

\subsection{6D rotation re-parameterization}\label{subsec:rotation_reparam}

We used for following parameterization: $R(x, y) = \operatorname{stack}[n(x), n(x \times y), n(x \times (x \times y))]$ where $n(x)$ is a normalization operation, and $a \times b$ is a vector product. This is essentially Gram-Schmidt orthogonalization. Rows of the resulting matrix is guaranteed to form an orthonormal basis. This rotation representation was shown to be better suited for optimization \cite{zhou2019continuity}.

\subsection{Stable Gaussian Mixture Likelihood}
We used the following numerically stable spherical GMM log-likelihood to compute \eq{ba_objective}:
\label{sec:gmm_stable_mixture}
\begin{gather*}
    \log \left[ \sum_r  w_r \cdot \frac{\exp(-\frac{1}{2}(\bm{x}-\bm{\mu}_r)^T \cdot (\sigma_r^2 \cdot I)^{-1} (\bm{x}-\bm{\mu}_r)}{\sqrt{(2\pi)^2 \sigma_r^4}} \right] \\ = \log \sum_r \exp \left[ \log\left(\frac{w_r}{2\pi\sigma_r^2} \right) - \frac{||\bm{x} - \bm{\mu}_r||^2}{2 \sigma_r^2} \right] \\
    = \operatorname{LSE}_r\left[\log\left(\frac{w_r}{2\pi\sigma_r^2} + \varepsilon \right) - \frac{||\bm{x} - \bm{\mu}_r||^2}{2 \sigma_r^2}\right]
\end{gather*}
where $(\mu, \sigma^2, w)$ are mean, variance and weight of the corresponding mixture component, $\operatorname{LSE}(l_0, \dots, l_r)$ is a numerically stable ``log-sum-exp'' often implemented as $\operatorname{LSE}(l_0, \dots, l_r) = l^* + \log(\sum_k \exp(l_k - l^*))$, where $l^* = \max(l_0, \dots, l_r)$, and $\varepsilon$ is a small number.

\subsection{Teacher loss}\label{supsec:teacher_loss}
In addition to the reprojcetion loss, the student-teacher ablation (S2/TS) used the following additional loss inspired by \citet{ma2020deepoptimizer} to draw the predicted solution $\bm{J}_{\text{neur}}$ into the basin of the correct solution by penalizing its deviation from the solution $\bm{J}_{\text{ref}}$ produced by the iterative refiner (IR). 
\begin{align}
    \mathcal L_{y}\big(\bm{J}_{\text{neur}}, & R^{(c)}_{\text{neur}}, \bm{t}^{(c)}_{\text{neur}}, s^{(c)}_{\text{neur}}; \bm{J}_{\text{ref}}, R^{(c)}_{\text{ref}}, \bm{t}^{(c)}_{\text{ref}}, s^{(c)}_{\text{ref}}\big) \nonumber \\
    = & \lambda_p \cdot ||\bm{J}_{\text{neur}} - \bm{J}_{\text{ref}}||_2^2 + \lambda_t \cdot \sum ||\bm{t}_{\text{neur}}^{(c)} - \bm{t}_{\text{ref}}^{(c)}||_2^2 \nonumber \\ 
    + & \lambda_R \cdot \sum ||(R_{\text{neur}}^{(c)})^T R_{\text{ref}}^{(c)} - I||_2^2 \nonumber \\
    + & \lambda_s \cdot \sum ||\log(s_{\text{neur}}^{(c)}) - \log(s_{\text{ref}}^{(c)})||_2^2.
\end{align}
Table \ref{tab:teacher_student} shows that it hurts the performance of the model.

\subsection{Qualitative Results}\label{sec:supp_images}
We provide qualitative examples (failure cases, success cases) on the test set of H36M and SkiPose in Figures \ref{fig:sup_example_h36m/1}-\ref{fig:sup_example_ski12/19}. Videos with more test prediction visualizations are available at: \href{https://bit.ly/cvpr22_6639}{https://bit.ly/cvpr22\_6639}. Circles around joints on 2D views represent the absolute \textit{reprojection error} for that joint for that view. Our qualitative findings:
\begin{enumerate}
    \item MetaPose considerably improves over the initial guess when a lot of self-occlusion is present
    \item MetaPose fails on extreme poses for which monocular estimation fails (e.g. somersaults)
    \item In two-camera SkiPose setup, AniPose often yields smaller reprojection error while producing very bad 3D pose results
\end{enumerate}

\clearpage
\input{figs/fign_examples}







%% file: tables/results_neural_abl.tex
\begin{table}[ht]
\small
\centering
\inctabcolsep{-3.3pt} {
\begin{tabular}{cccccccccccccccc}\toprule
$\joints,\cameras$ &\cmark &\cmark &\cmark &\cmark &\xmark &\xmark &\xmark &\xmark &\cmark &\cmark &\cmark &\xmark &\cmark & \xmark &\xmark \\
 $\heatmaps$ & \cmark & \cmark & \cmark & \xmark & \cmark & \xmark & \cmark & \cmark & \xmark & \xmark & \cmark & \xmark & \xmark & \cmark & \xmark \\
 $\keypoints$ & \cmark & \cmark & \xmark & \cmark & \cmark & \cmark & \xmark & \cmark & \xmark & \cmark & \xmark & \cmark & \xmark & \xmark & \xmark \\
 $\mathcal L$ & \cmark & \xmark & \cmark & \cmark & \cmark & \cmark & \cmark & \xmark & \cmark & \xmark & \xmark & \xmark & \xmark & \xmark & \cmark \\ \midrule
 \textbf{4} & \textbf{32} & 33 & \textbf{32} & 45 & \textbf{32} & 48 & 33 & \textbf{32} & 45 & 45 & \textbf{32} & 46 & 45 & 34 & nan \\
 \textbf{3} & {36} & \textbf{35} & {36} & 49 & 37 & 50 & 38 & 39 & 49 & 50 & 37 & 50 & 49 & 44 & nan \\
 \textbf{2} & \textbf{44} & \textbf{44} & 46 & 53 & 43 & 59 & 40 & 56 & 53 & 58 & 45 & 55 & 51 & 45 & nan \\ \bottomrule
\end{tabular}
}
\caption{
\textbf{Ablation} --
PMPJPE$\downarrow$ of our method on Human3.6M with different number of cameras with different inputs passed to the neural optimizer. Heatmap $\heatmaps$ contributes most to the final performance, but all inputs are necessary to achieve the state-of-art performance. \label{tab:neural_abl}
\vspace{-10px}
}
\end{table}

%% file: tables/sup_latency_breakdown.tex
\begin{table*}[]
\centering
\begin{tabular}{lcccccc}\toprule
 & \textbf{PoseNet} & \textbf{GMM} & \textbf{S1} & \textbf{Solver} & \textbf{Total} & \textbf{Error {[}mm{]}} \\ \midrule
AniPose & 0.03 $\cdot$ 4 & - & - & 7 & 7.1 & 75 \\
MetaPose (S1) & 0.03 $\cdot$ 4 & - & 0.01 $\cdot$ 4 & - & \textbf{0.15} & 74 \\
MetaPose (S1+S2) & 0.03 $\cdot$ 4 & 0.01 $\cdot$ 4 & 0.01 $\cdot$ 4 & 0.006 & 0.2 & \textbf{40} \\
MetaPose (S1+IR) & 0.03 $\cdot$ 4 & 0.01 $\cdot$ 4 & 0.01 $\cdot$ 4 & 1.5 & 1.7 & 43 \\
\midrule
AniPose & 0.5 $\cdot$ 4 & - & - & 10 & 12 & 75 \\
MetaPose (S1) & 0.5 $\cdot$ 4 & - & 0.20 $\cdot$ 4 & - & \textbf{2.8} & 74 \\
MetaPose (S1+S2) & 0.5 $\cdot$ 4 & 0.25 $\cdot$ 4 & 0.20 $\cdot$ 4 & 0.01 & 4 & \textbf{40} \\
MetaPose (S1+IR) & 0.5 $\cdot$ 4 & 0.25 $\cdot$ 4 & 0.20 $\cdot$ 4 & 3.5 & 7.5 & 43 \\
\bottomrule
\end{tabular}
\caption{Latency breakdown in seconds for estimating the full 3D pose on H36M with four cameras on a GPU (V100, top) and a CPU (bottom) across four components: per-view 2D heatmap estimation (PoseNet), heatmap GMM fitting, per-view monocular 3D and initialization, multi-view bundle adjustment (neural network forward pass in case of MetaPose S1+S2, Adam \cite{kingma2014adam} in case of S1+IR, and a 2nd-order CPU-only TRR \cite{byrd1988approximate,branch1999subspace} solver in case of AniPose). MetaPose (S1+S2) achieves lowest error with an at least six times (on GPU; two times on CPU) faster inference as the iterative refiner. \label{tab:latency}}
\end{table*}

%% file: tables/sup_mlp_ski_h36m.tex
\begin{table}[p]
\centering
\begin{tabular}{lccc}
\multicolumn{3}{c}{(a) H36M} \\ \toprule
\textbf{Method} & \textbf{4} & \textbf{3} & \textbf{2} \\ \midrule
Metapose (S1+S2) & 32 & 36 & 44 \\
Metapose (S1+S2/MLP) & \textbf{30} & \textbf{35} & \textbf{41} \\ 
\bottomrule \\
\multicolumn{3}{c}{(b) SkiPose} \\ \toprule
\textbf{Method} & \textbf{6} & \textbf{4} & \textbf{2} \\ \midrule
Metapose (S1+S2) & \textbf{42} & \textbf{45} & \textbf{50} \\
Metapose (S1+S2/MLP) & 46 & 44 & 54 \\ \bottomrule 
\end{tabular}
\caption{Equivariant (S1+S2) and non-equivariant (S1+S2/MLP) performance networks have comparable performance across different numbers of cameras on H36M (top) and SkiPose (bottom).}\label{tab:skipose_mlp}
\end{table}

%% file: tables/sup_results_h36m_ski_extended.tex
\begin{table}[t]
\begin{center}
\inctabcolsep{-3pt} {
\begin{tabular}{lc|cccc|c}
\multicolumn{7}{c}{(a) H36M} \\ \toprule
\multicolumn{2}{l}{\multirow{2}{*}{Method and supervision type}} & 
\multicolumn{2}{c}{PMPJPE$\downarrow$} & \multicolumn{2}{c}{NMPJPE$\downarrow$} &  \multirow{2}{*}{\stackanchor{$\Delta t$}{[s]}} \\
\cmidrule(lr){3-4} \cmidrule(lr){5-6}
& & 4 & 2 & 4 & 2 \\
\midrule
\rowcolor[gray]{0.9}
Isakov et al. \cite{iskakov2019learnable} & \textbf{3D} & 20 & - & - & - & - \\ 
\midrule
AniPose \cite{karashchuk2020anipose} w/ GT & S & 75 & 167 & 103 & 230 & 7.1 \\
Rhodin et al. \cite{rhodin2018skipose} & 2/3D & 65 & - & 80 & - & - \\
CanonPose \cite{wandt2020canonpose} & S & 53 & - & 82 & - & - \\
EpipolarPose (EP) \cite{kocabas2019epipolar} & S & 71 & - & 78 & - & - \\
Iqbal et al. \cite{iqbal2020weakly} & 2D & 55 & - & 66 & - & - \\
\midrule
\textbf{MetaPose} (S1) & S & 74 & 87 & 83 & 95 & 0.2 \\
\textbf{MetaPose} (S1+S2) & 2D & \textbf{32} & \textbf{44} & \textbf{49} & \textbf{55} & 0.2 \\
\textbf{MetaPose} (S1+IR) & S & 43 & 66 & 53 & 75 & 1.7 \\
\textbf{MetaPose} (S1+S2/SS) & S & 39 & 50 & 56 & 63 & 0.2 \\
\bottomrule
\end{tabular}

\bigskip
\begin{tabular}{lc|cccc|c}
\multicolumn{7}{c}{(b) SkiPose} \\ \toprule
\multicolumn{2}{l}{\multirow{2}{*}{Method and supervision type}} & 
\multicolumn{2}{c}{PMPJPE$\downarrow$} & \multicolumn{2}{c}{NMPJPE$\downarrow$} &  \multirow{2}{*}{\stackanchor{$\Delta t$}{[s]}} \\
\cmidrule(lr){3-4} \cmidrule(lr){5-6}
& & 6 & 2 & 6 & 2 \\
\midrule
AniPose \cite{karashchuk2020anipose} w/ GT & S & 50 & 62 & 221 & 273 & 7.1 \\
Rhodin et al. \cite{rhodin2018skipose} & 2/3D & - & - & 85 & - & - \\
CanonPose (CP) \cite{wandt2020canonpose} & S & 90 & - & 128 & - & - \\
\midrule
\textbf{MetaPose} (S1) & S & 81 & 86 & 140 & 144 & 0.3 \\
\textbf{MetaPose} (S1+S2) & 2D & 42 & \textbf{50} & \textbf{53} & \textbf{59} & 0.4 \\
\textbf{MetaPose} (S1+IR) & S & \textbf{30} & 77 & 54 & 94 & 2.5 \\
\textbf{MetaPose} (S1+S2/SS) & S & 42 & 95 & 59 & 102 & 0.4 \\
\bottomrule
\end{tabular}
}
\end{center}
\vspace{-1em}
\caption{
\textbf{MetaPose outperforms SotA on H36M (top) and SkiPose (bottom)} --
Same notation as in Table~\ref{tab:quantitative}. Also includes the self-supervised (S2/SS) and iterative solver (SS/IR) flavours of MetaPose. Supervision signal used \textit{during training}: 2D - ground truth 2D keypoints, 3D - ground truth 3D poses, S - self-supervision (i.e. using a pose estimation network pre-trained on a different dataset), 2/3D - 2D keypoint data with 3D poses on few subjects. 
}
\label{tab:quantitative_extended}
\end{table}

%% file: tables/sup_bone_lengths.tex

\begin{table*}[ht]
\centering
\begin{tabular}{lccc} 
\multicolumn{4}{c}{(a) H36M} \\
\toprule
\textbf{Method} & \textbf{4} & \textbf{3} & \textbf{2} \\ \midrule
Metapose S1+S2 & 32 & 36 & 44 \\
Metapose S1+S2 \textbf{+ bone} & \textbf{31} & \textbf{34} & \textbf{37} \\ \midrule
Metapose S1+IR & 43 & 52 & 53 \\
Metapose S1+IR \textbf{+ bone} & \textbf{38} & \textbf{44} & \textbf{47} \\ \midrule
Metapose S1+S2/SS & 39 & 47 & \textbf{50} \\
Metapose S1+S2/SS \textbf{+ bone} & \textbf{38} & \textbf{45} & \textbf{50} \\ \bottomrule \\ 
\multicolumn{4}{c}{(b) SkiPose} \\
\toprule
\textbf{Method} & \textbf{6} & \textbf{4} & \textbf{2} \\ \midrule
Metapose S1+S2 & \textbf{41} & \textbf{43} & \textbf{47} \\
Metapose S1+S2 \textbf{+ bone} & 45 & 46 & 49 \\ \midrule 
Metapose S1+IR & 30 & 32 & 77 \\
Metapose S1+IR \textbf{+ bone} & \textbf{26} & \textbf{28} & \textbf{46} \\ \midrule
Metapose S1+S2/SS & 41 & 46 & 95 \\
Metapose S1+S2/SS \textbf{+ bone} & \textbf{44} & \textbf{45} & \textbf{53} \\
\bottomrule
\end{tabular}
\bigskip \bigskip
\caption{Personalized \textbf{bone lengths prior} helps in all cases for H36M (top), especially in the few-camera setup; and in the majority of cases on SkiPose (bottom). \label{tab:bone_lengths}}
\end{table*}

%% file: tables/sup_teacher_student.tex
\begin{table*}[ht]
\centering
\begin{tabular}{lccc} \multicolumn{4}{c}{(a) H36M} \\ \toprule
\textbf{Method} & \textbf{4} & \textbf{3} & \textbf{2} \\
\midrule
MetaPose S1+S2 & \textbf{32} & \textbf{36} & \textbf{44} \\ 
MetaPose S1+S2/TS & 38 & 45 & 45 \\
\bottomrule \\
\multicolumn{4}{c}{(b) SkiPose} \\ \toprule
\textbf{Method} & \textbf{6} & \textbf{4} & \textbf{2} \\
\midrule
MetaPose S1+S2 & \textbf{42} & 45 & \textbf{50} \\
MetaPose S1+S2/TS & \textbf{42} & \textbf{43} & 72 \\
\bottomrule
\end{tabular}
\bigskip \bigskip
\caption{\textbf{Teacher-student loss} analogous to the one proposed by \citet{ma2020deepoptimizer} to bring the neural optimizer into the basin of the correct solution either hurts or does significantly affect the performance in all cases. \label{tab:teacher_student}}
\end{table*}

%% file: tables/sup_monocular_details.tex
\begin{table*}
\begin{center}
\begin{tabular}{lccccccccc}\toprule
Metric & \multicolumn{3}{c}{\textbf{Train}} &  \multicolumn{3}{c}{\textbf{Validation}} &  \multicolumn{3}{c}{\textbf{Test}} \\ \midrule
GT log-prob. &  \multicolumn{3}{c}{-5.17} & \multicolumn{3}{c}{-5.58} & \multicolumn{3}{c}{-5.06} \\
\cmidrule(lr){2-4} \cmidrule(lr){5-7} \cmidrule(lr){8-10} 
Stage: & S1 & S1+IR & S1+S2 & S1 & S1+IR & S1+S2 & S1 & S1+IR & S1+S2 \\ \midrule
Pred log-prob. &  -4.22 & -5.86 & -5.00 & -4.6 & -6.07 & -5.41 & -3.92 & -5.77 & -4.95 \\
PMPJPE [mm] & 69 & 38 & 15 & 65 & 34 & 17 & 74 & 43 & 32 \\
NMPJPE [mm] & 78 & 58 & 36 & 69 & 56 & 46 & 88 & 66 & 49 \\
MSE 2D ($10^{-4}$) & 15 & 5 & 0.6 & 6 & 2 & 0.6 & 20 & 7 & 5  \\ \bottomrule
\end{tabular}

\smallskip (a) H36M
\bigskip

\begin{tabular}{lccccccccc}\toprule
Metric & \multicolumn{3}{c}{\textbf{Train}} &  \multicolumn{3}{c}{\textbf{Validation}} &  \multicolumn{3}{c}{\textbf{Test}} \\ \midrule
GT log-prob. &  \multicolumn{3}{c}{-5.49} & \multicolumn{3}{c}{-5.49} & \multicolumn{3}{c}{-4.81} \\
\cmidrule(lr){2-4} \cmidrule(lr){5-7} \cmidrule(lr){8-10} 
Stage: & S1 & S1+IR & S1+S2 & S1 & S1+IR & S1+S2 & S1 & S1+IR & S1+S2 \\ \midrule
Pred log-prob. &  -2.83 & -5.79 & -5.49 & -2.90 & -5.75 & -4.51 & -3.04 & -5.59 & -5.33 \\
PMPJPE [mm] & 71 & 17 & 1 & 72 & 17 & 10 & 80 & 30 & 42 \\
NMPJPE [mm] & 139 & 35 & 1 & 143 & 38 & 15 & 140 & 54 & 53 \\
MSE 2D ($10^{-4}$) & 34 & 6 & 0.01 & 37 & 6 & 1 & 30 & 7 & 7  \\ \bottomrule
\end{tabular}

\smallskip (b) SkiPose \bigskip

\caption{\textbf{Details about predictions across different stages: initialization using monocular 3d (S1), iterative refinement (S1+IR), and neural refinement (S1+S2) on H36M with four cams (top) and SkiPose with six cams (bottom)}. 2D error is scaled so that the entire pose lies in $[0, 1]^2$. The GT log probability is the log probability of ground truth points given predicted heatmaps and measures how well heatmaps generated by our 2D prediction network match the ground truth. Significantly larger discrepancy between GT log probabilities on train and test on SkiPose shows that 2D pose network overfits much more on SkiPose than on H36M due to its limited size.}\label{tab:mono_data}

\end{center}
\end{table*}

%% file: tables/sup_camera_model.tex
\begin{table*}[th]
\centering
\begin{tabular}{lccccc} 
\multicolumn{6}{c}{(a) H36M} \\ \toprule
\textbf{Method} & \textbf{2D} & \textbf{S1} & \textbf{4} & \textbf{3} & \textbf{2} \\ \midrule
S1+IR & HT & EP & 43 & 52 & 53 \\
S1+IR & HT & GT & 40 & 49 & 48 \\
S1+IR & full-GT & EP & 17 & 20 & 24 \\
S1+IR & full-GT & GT & 14 & 16 & 20 \\
S1+IR & weak-GT & EP & 4 & 6 & 18 \\
S1+IR & weak-GT & GT & 1.4 & 1.7 & 2 \\ \bottomrule \\
\multicolumn{6}{c}{(b) SkiPose} \\ \toprule
\textbf{Method} & \textbf{2D} & \textbf{S1} & \textbf{6} & \textbf{4} & \textbf{2} \\ \midrule
S1+IR & HT & CP & 30 & 33 & 77 \\
S1+IR & HT & GT & 28 & 30 & 41 \\
S1+IR & full-GT & CP & 8 & 8 & 29 \\
S1+IR & weak-GT & CP & 8 & 7 & 28 \\ \bottomrule
\end{tabular}
\bigskip
\caption{MetaPose S1+IR trained with either ground truth pseudo-heatmaps centered around full and \textbf{weak-projected 3D joints} and with different S1 \textbf{initialization} (either predicted via EpipolarPose or ``perfect''). This experiment shows that imperfect heatmaps contribute to at least 20mm of error in both cases, weak camera model contribute to 10mm of error on H36M and no error on SkiPose, and imperfect initialization contributes to at most 3mm of error.  \label{tab:camera_model}}
\end{table*}

%% file: tables/sup_h36m_size.tex
\begin{table*}
\begin{center}
\inctabcolsep{-2pt} {
\begin{tabular}{ccccccccccccccccccccccccc}\toprule
\backslashbox[20mm]{\textbf{\# cam}}{\textbf{\ \% \ }} & \textbf{100} & \textbf{89} & \textbf{84} & \textbf{79} & \textbf{73} & \textbf{68} & \textbf{63} & \textbf{58} & \textbf{52} & \textbf{47} & \textbf{42} & \textbf{37} & \textbf{31} & \textbf{29} & \textbf{26} & \textbf{24} & \textbf{21} & \textbf{18} & \textbf{16} & \textbf{13} & \textbf{10} & \textbf{8} & \textbf{5} & \textbf{3} \\ \midrule
\textbf{4} & \textbf{32} & \textbf{32} & \textbf{32} & \textbf{32} & 33 & 36 & 33 & 35 & 33 & 35 & 34 & 42 & 37 & 39 & 36 & 38 & 36 & 41 & 48 & 41 & 41 & 44 & 48 & 70 \\
\textbf{3} & 36 & \textbf{35} & 36 & 37 & 37 & 36 & 37 & 39 & 37 & 37 & 38 & 39 & 40 & 39 & 41 & 45 & 42 & 43 & 44 & 45 & 46 & 51 & 53 & 70 \\
\textbf{2} & 44 & 48 & 48 & 47 & 46 & 48 & 48 & \textbf{40} & 54 & 60 & 43 & 43 & 51 & 57 & 53 & 58 & 54 & 50 & 52 & 48 & 48 & 51 & 68 & 87 \\
\bottomrule
\end{tabular}
}
\bigskip
\end{center}
\caption{\textbf{Test PMPJE of MetaPose on H36M as a function of the fraction of training examples with 2D ground truth used (i.e. first X\%)}. Reminder: we \textbf{never} use any ground truth 3D annotations for either cameras or poses, these are percentages of 2D labels used for training. We can see that MetaPose produces high-accuracy predictions (within 10mm of the original performance) with up to 1/5-th ($\approx$18\%) of the H36M training 2D pose annotations ($\approx$5k training examples each containing multiple cameras). The few-camera setup exhibits more variations in test error due to random network initialization.}
\label{tab:sup_h36m_size}
\end{table*}



%% file: tables/sup_gmm.tex
\begin{table*}[ht] \centering
\begin{tabular}{lcccc}
\multicolumn{5}{c}{(a) H36M} \\ \toprule
\textbf{Method} & \textbf{GMM} & \textbf{4} & \textbf{3} & \textbf{2} \\ 
\midrule
MetaPose S1+IR & 4 & 43 & 52 & 53 \\
MetaPose S1+IR & 3 & 42 & 51 & 52 \\
MetaPose S1+IR & 2 & 42 & 51 & 52 \\
MetaPose S1+IR & 1 & 42 & 52 & 53 \\
MetaPose S1+S2 & 4 & 32 & 39 & 44 \\
MetaPose S1+S2 & 3 & 31 & 36 & 47 \\
MetaPose S1+S2 & 2 & 32 & 36 & 50 \\
MetaPose S1+S2 & 1 & 32 & 36 & 48 \\
\bottomrule \\
\multicolumn{5}{c}{(b) SkiPose} \\ \toprule
\textbf{Method} & \textbf{GMM} & \textbf{6} & \textbf{4} & \textbf{2} \\ \midrule
MetaPose S1+IR & 4 & 30 & 33 & 77 \\
MetaPose S1+IR & 3 & 30 & 32 & 77 \\
MetaPose S1+IR & 2 & 31 & 34 & 75 \\
MetaPose S1+IR & 1 & 43 & 43 & 58 \\
MetaPose S1+S2 & 4 & 42 & 45 & 50 \\
MetaPose S1+S2 & 3 & 44 & 41 & 50 \\
MetaPose S1+S2 & 2 & 42 & 49 & 51 \\
MetaPose S1+S2 & 1 & 41 & 43 & 47 \\ 
\bottomrule
\end{tabular}
\bigskip \bigskip
\caption{The number of Gaussian Mixture components does not significantly affect the performance of the network in all cases on both SkiPose (top) and H36M (bottom), except for MetaPose S1+IR on SkiPose with a single Gaussian. \label{tab:gmm}}
\end{table*}

%% file: figs/fign_labl_fmt.tex
\begin{figure}
\begin{center}
\includegraphics[width=\linewidth,trim=0 4in 9in 0]{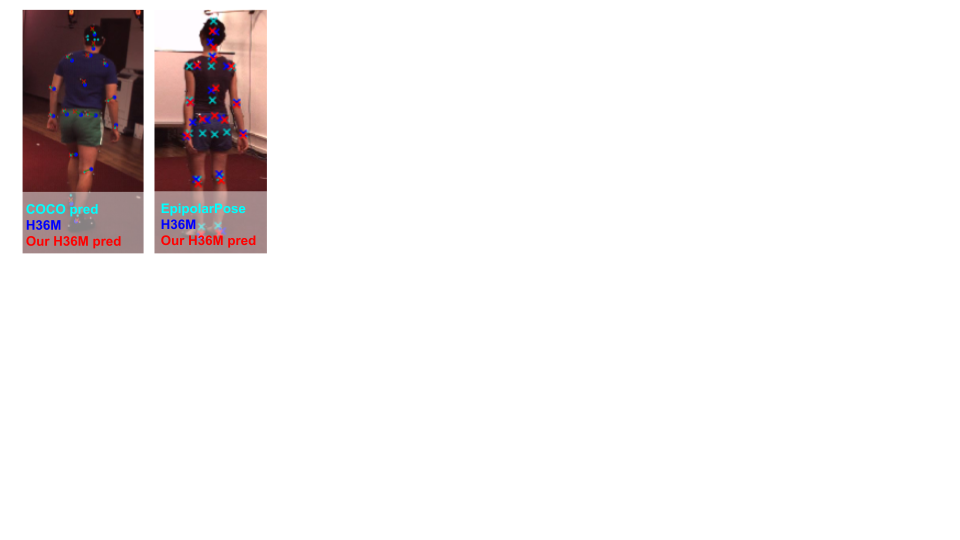}
\end{center}
\caption{Both H36M ground truth poses, COCO dataset (used to train the hourglass network), and EpipolarPose predictions (used to generate the 3D initialization) have different label formats from H36M. We trained a small ``adapter'' to convert COCO-to-H36M, and used EpipolarPose predictions as-is.}
\label{fig:fig_label_formats}
\end{figure}

%% file: figs/fign_examples.tex
\newcommand{\exfignew}[2]{
    \begin{figure*}
    \begin{center}
    \includegraphics[width=\linewidth]{imgs/sup/examples/#1.png}
    \end{center}
    \caption{#2}\label{fig:sup_example_#1}
    \end{figure*}
}

\newcommand{\exfignewcrop}[2]{
    \begin{figure*}
    \begin{center}
    \includegraphics[width=0.4\linewidth,trim=0 0in 23.8in 0,clip]{imgs/sup/examples/#1.png}
    \includegraphics[width=0.4\linewidth,trim=24in 0 0in 0,clip]{imgs/sup/examples/#1.png}
    \end{center}
    \caption{#2}\label{fig:sup_example_#1}
    \end{figure*}
}

\exfignew{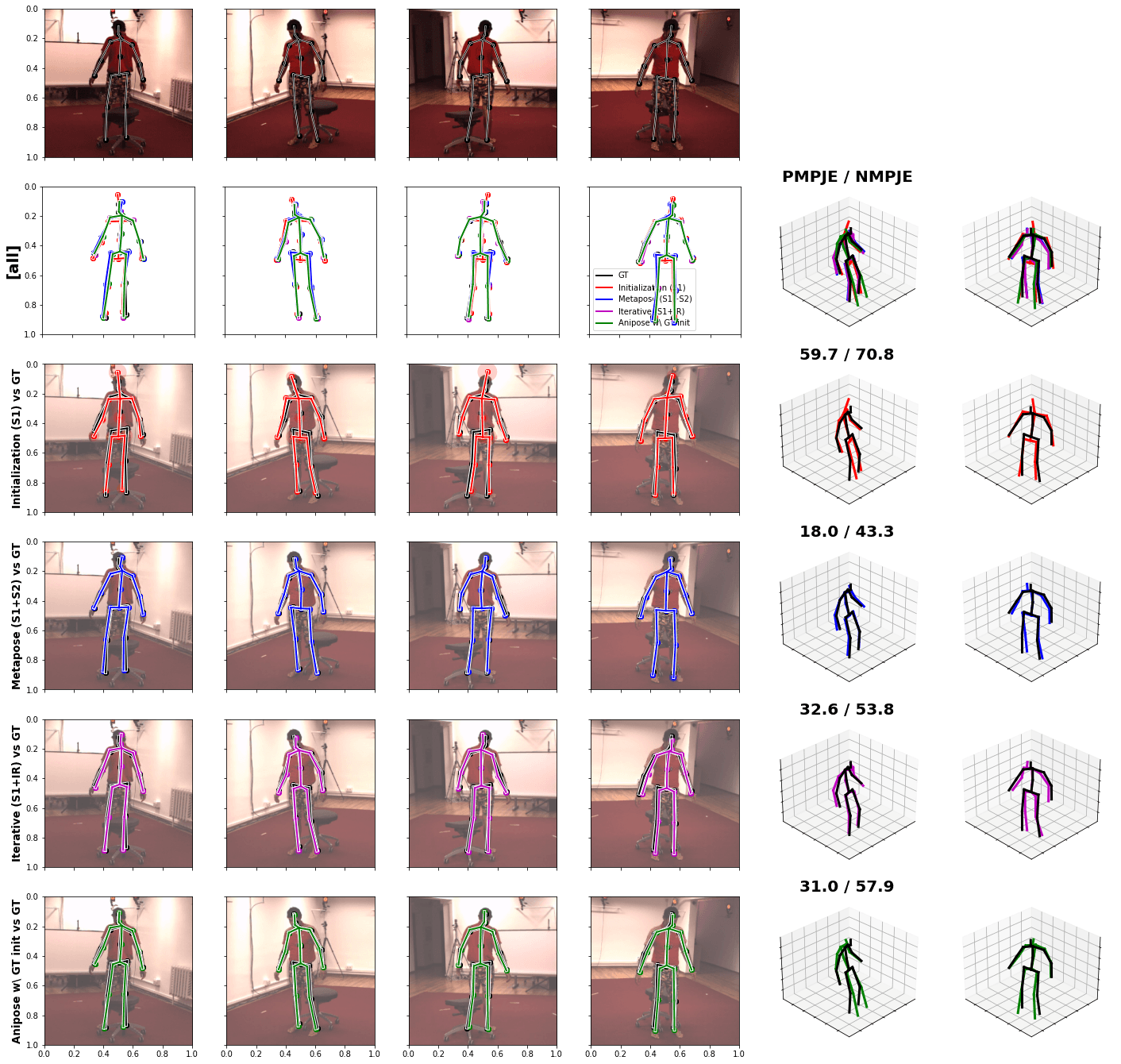}{Full MetaPose (S1+S2) outperforms initialization (S1), Iterative Solver (S1+IR), and AniPose w/ GT camera init.}

\exfignew{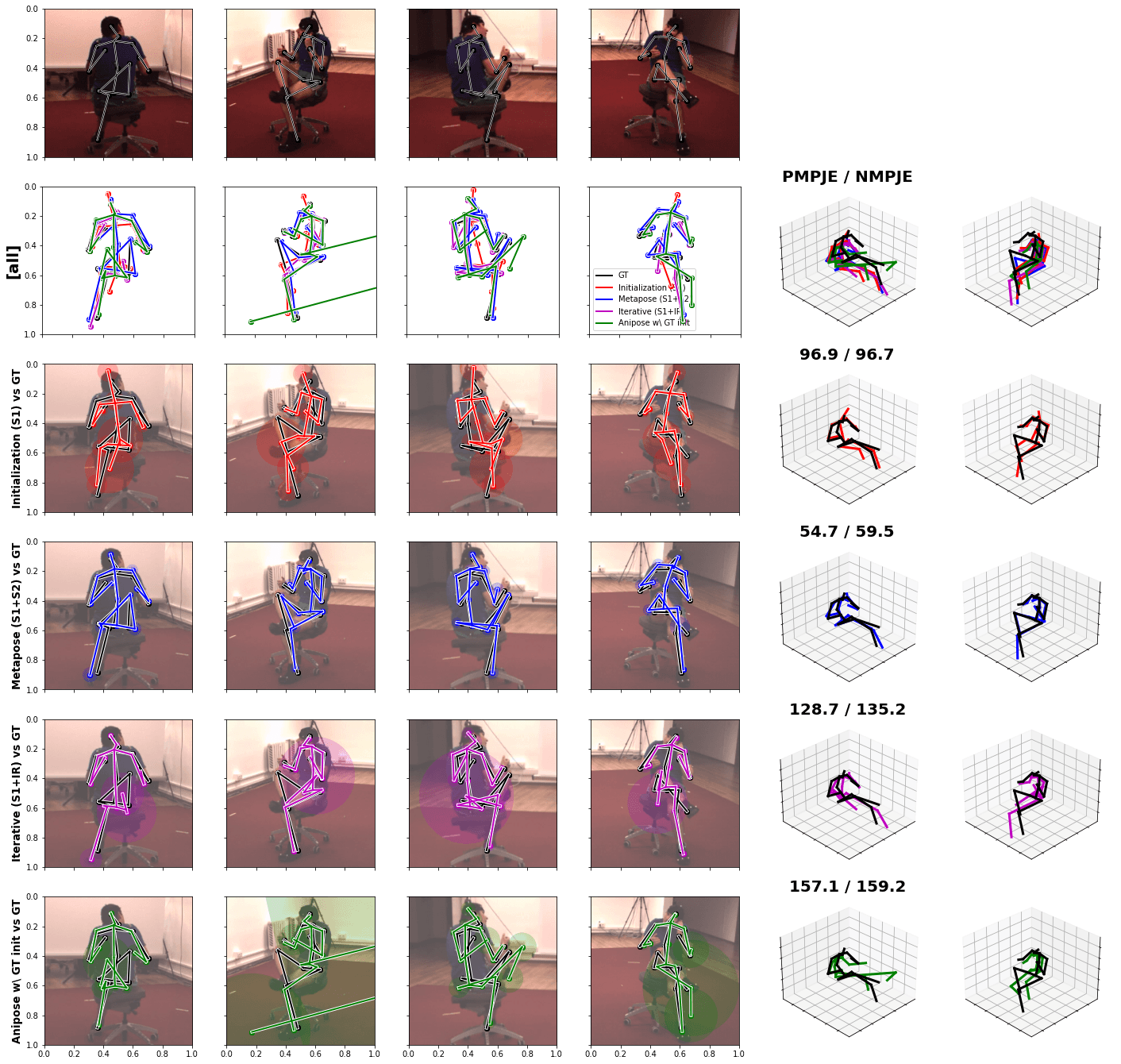}{MetaPose improves over the initial guess under high self-occlusion.}
\exfignew{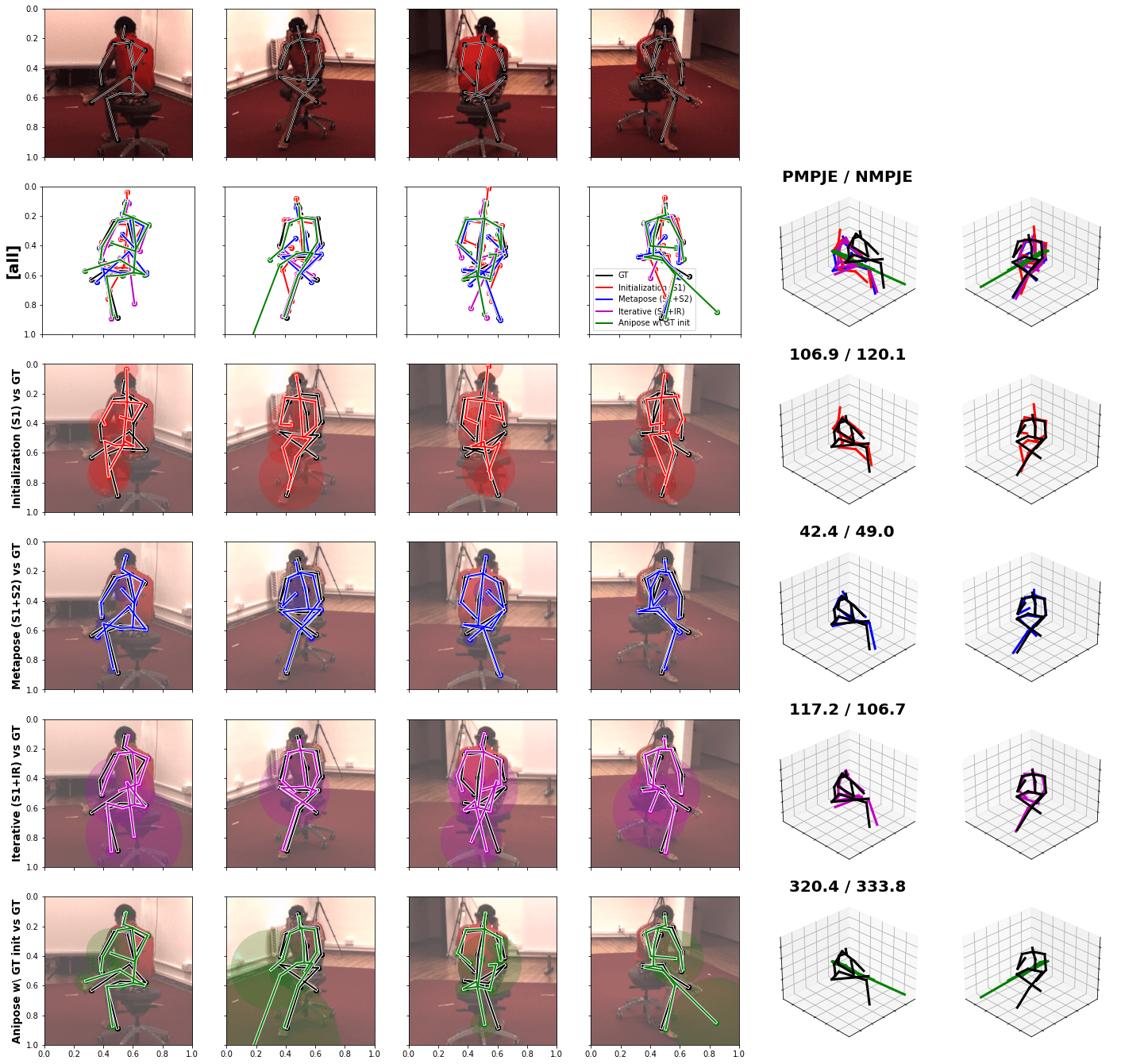}{MetaPose improves over the initial guess under high self-occlusion.}

\exfignew{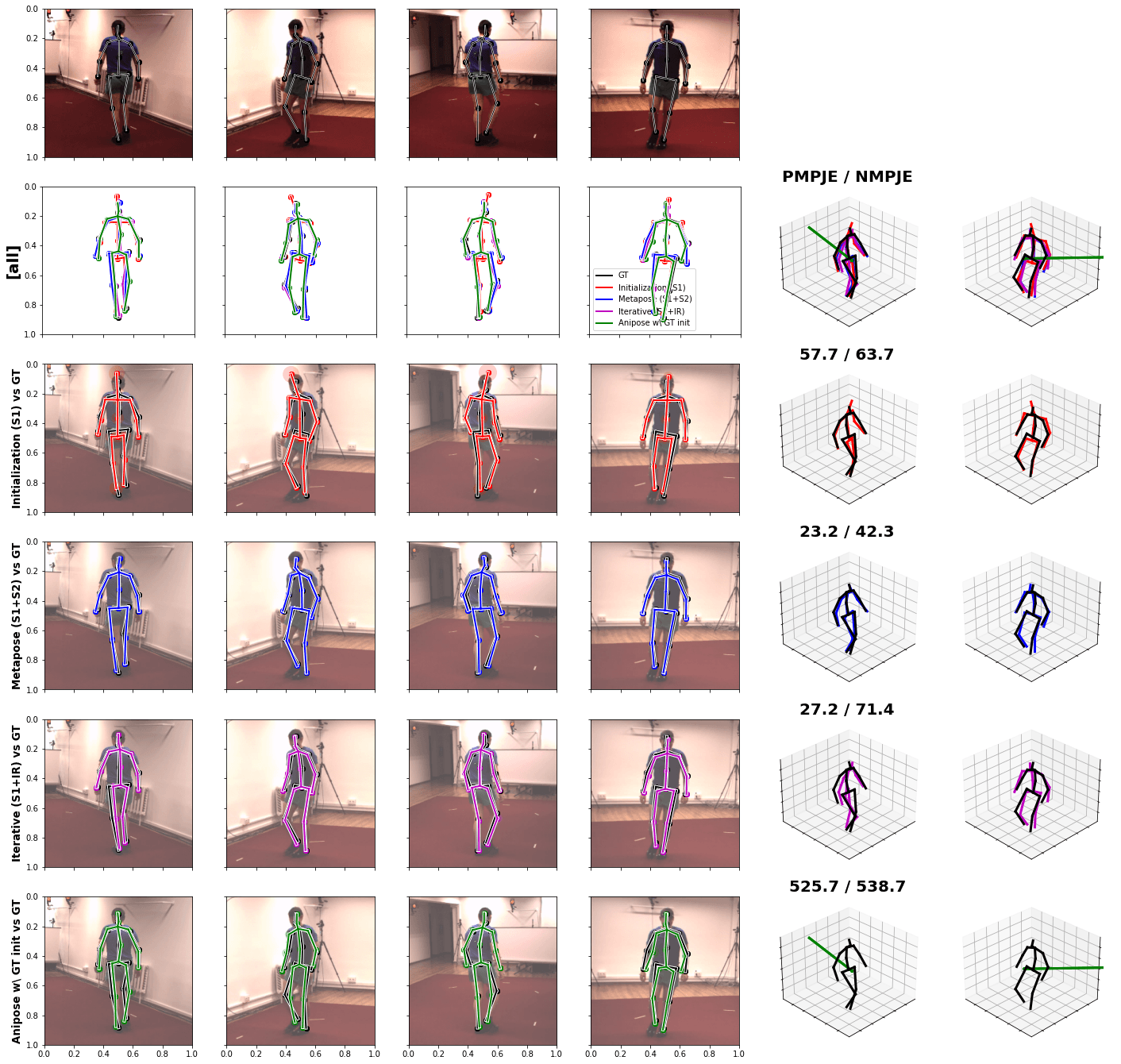}{AniPose w/ GT camera initialization can yields low re-projection error but high 3D estimation error.}

\exfignew{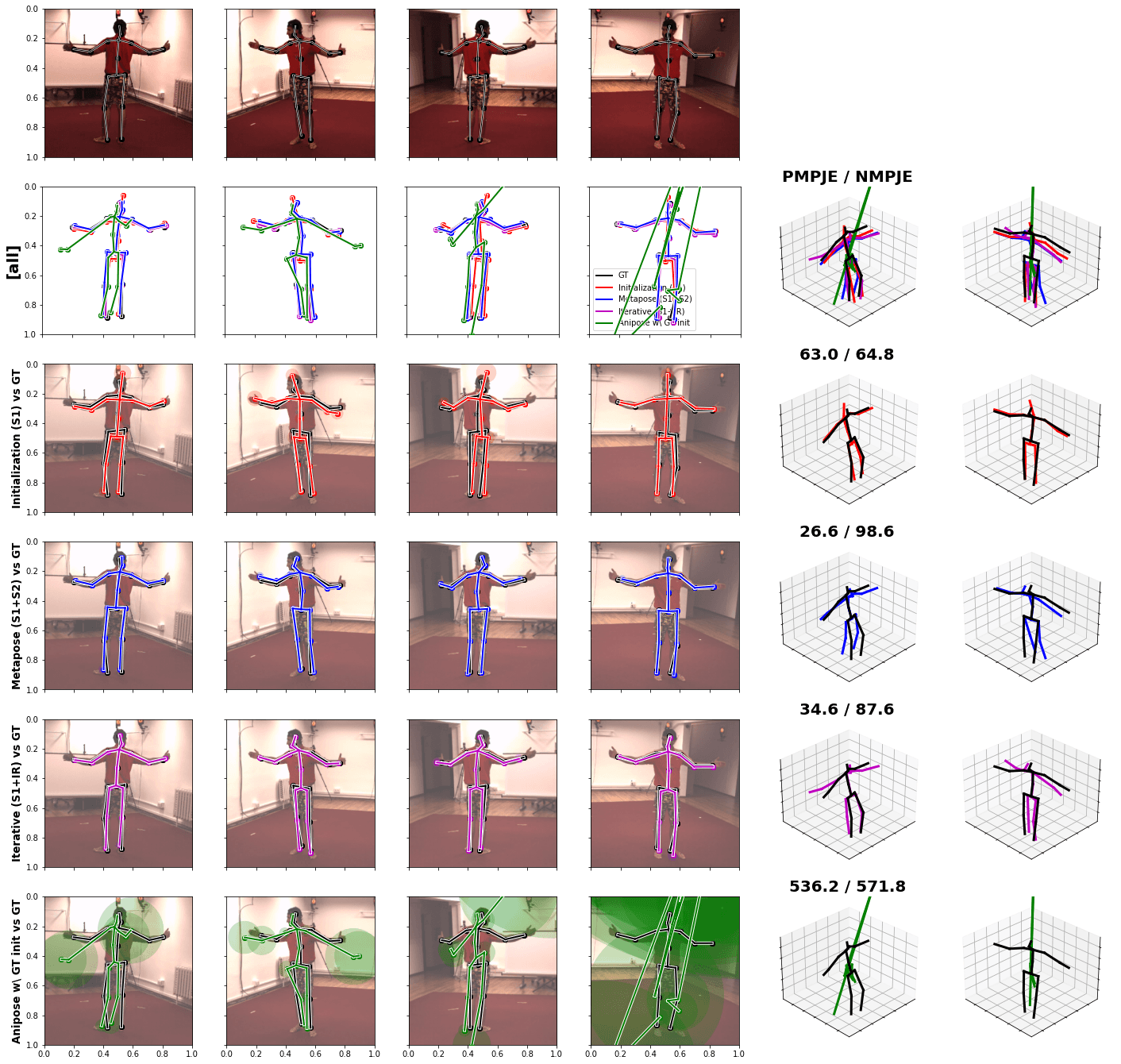}{AniPose with GT init fails due to poor choice of 2D predictions to ignore during refinement.}
\exfignew{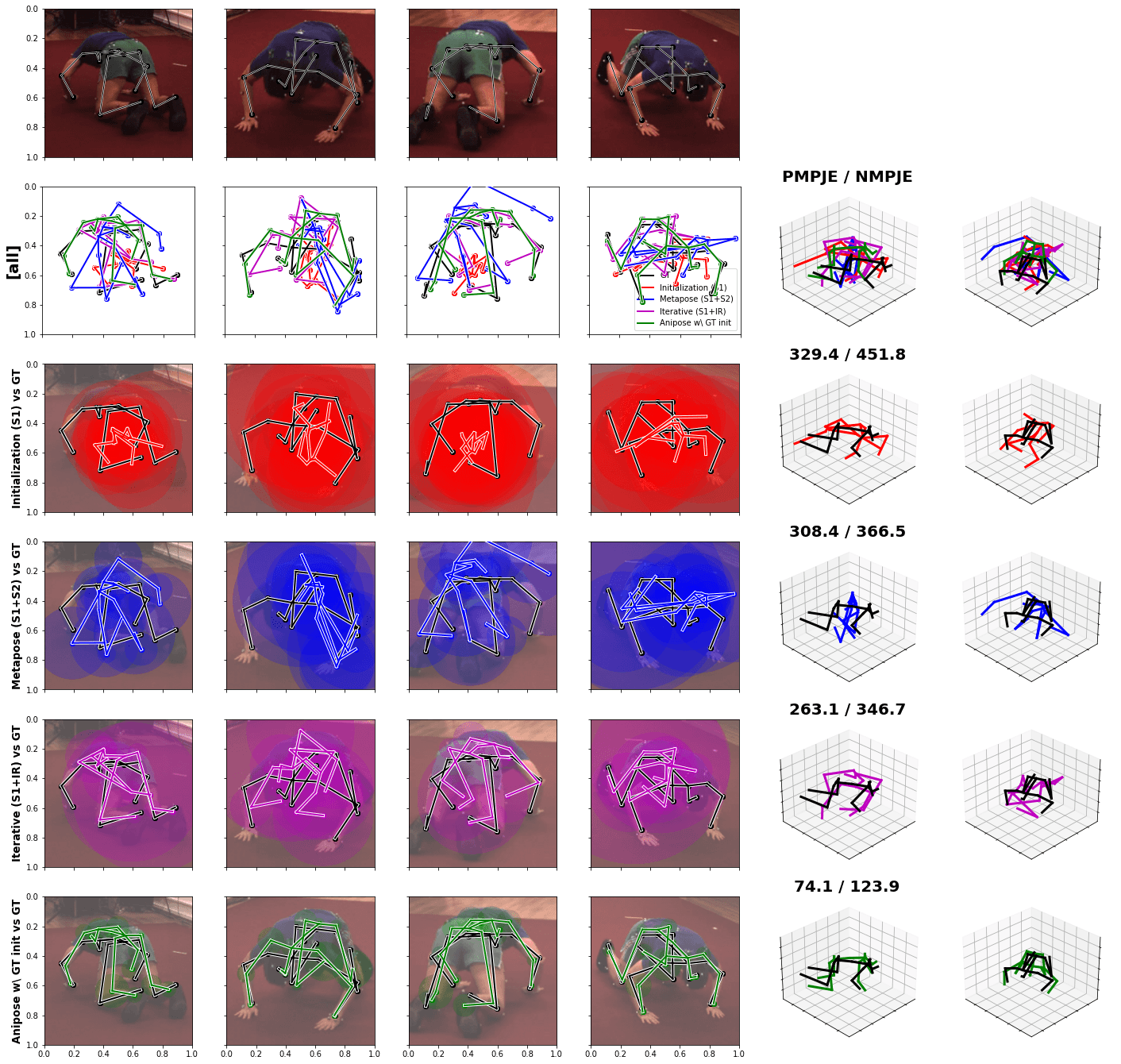}{MetaPose fails on few extreme poses that have much poorer than average initialization quality.}

\exfignew{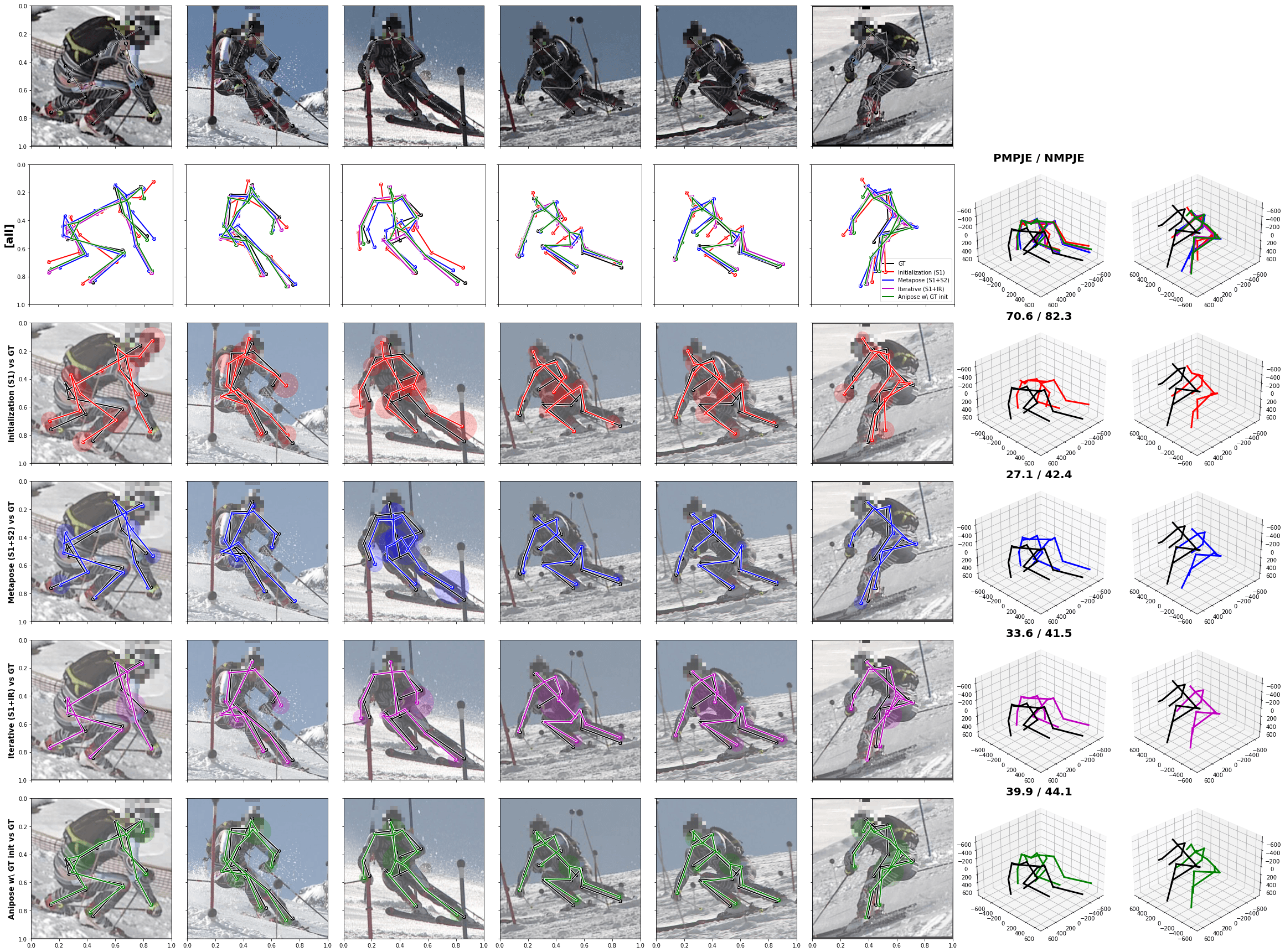}{MetaPose improves over the initial guess under high self-occlusion.}

\exfignew{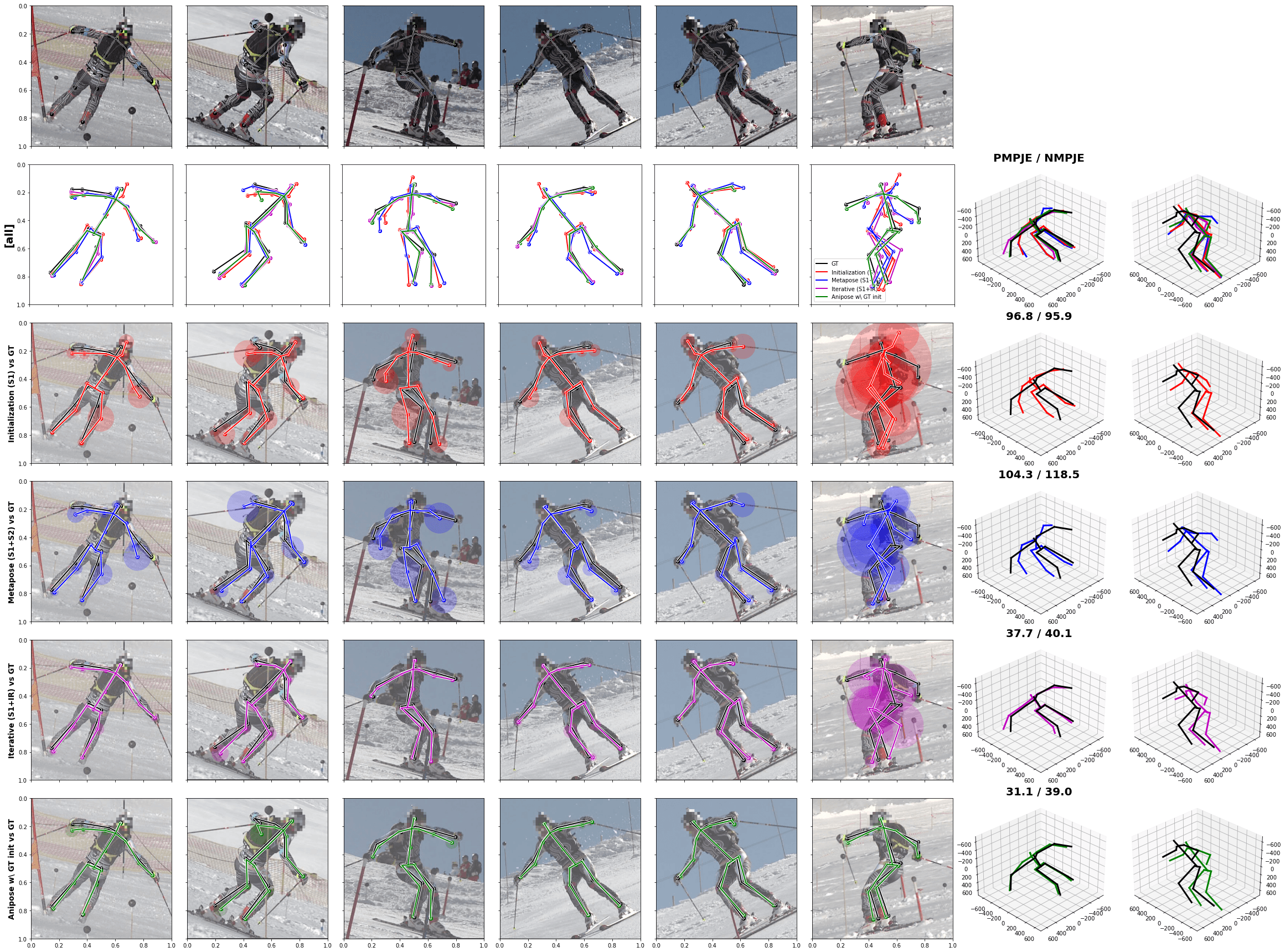}{MetaPose fails on poses that have much poorer (than average) initialization quality.}

\exfignew{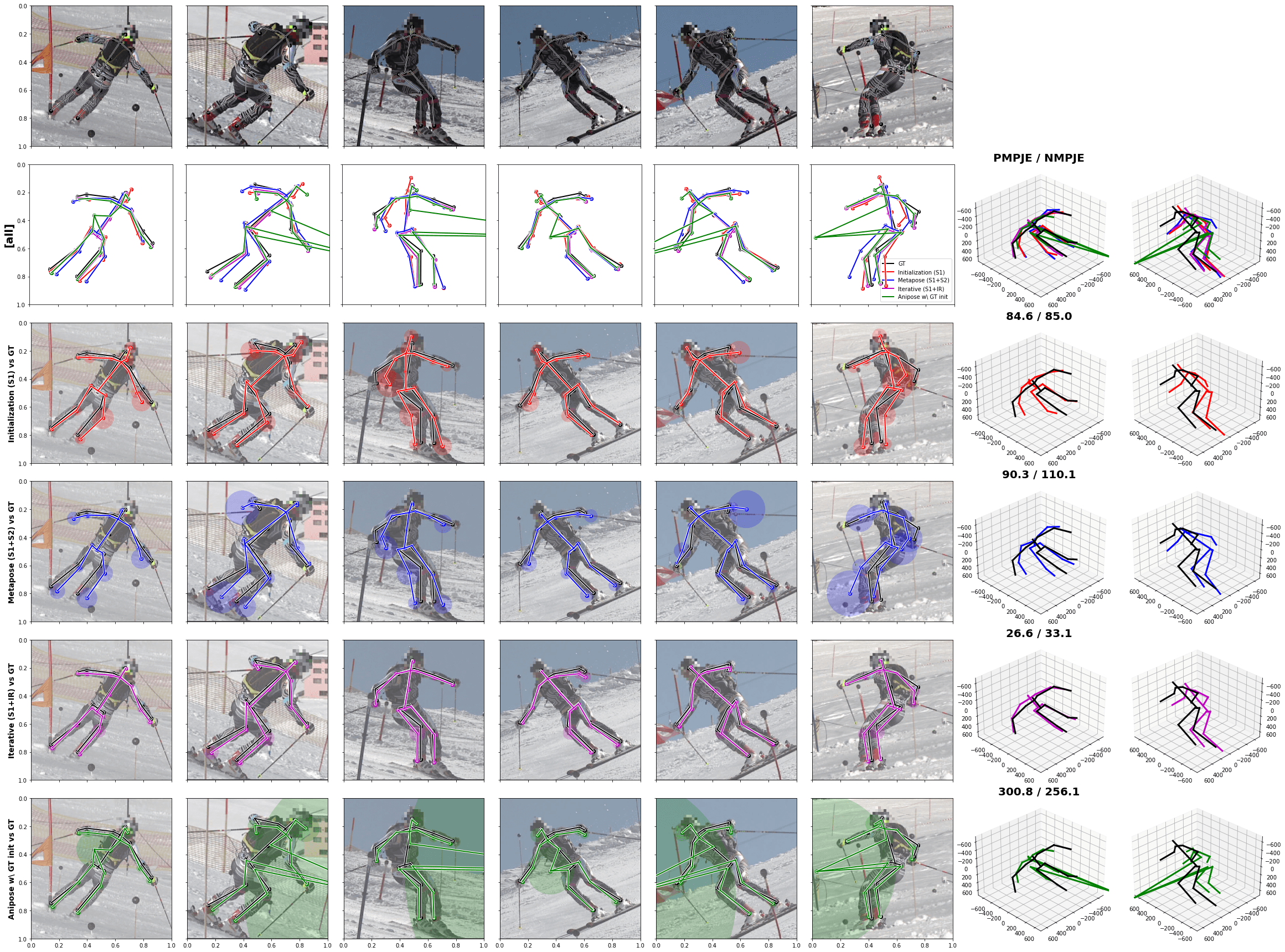}{AniPose with GT init fails due to poor choice of 2D predictions to ignore during refinement.}

\exfignewcrop{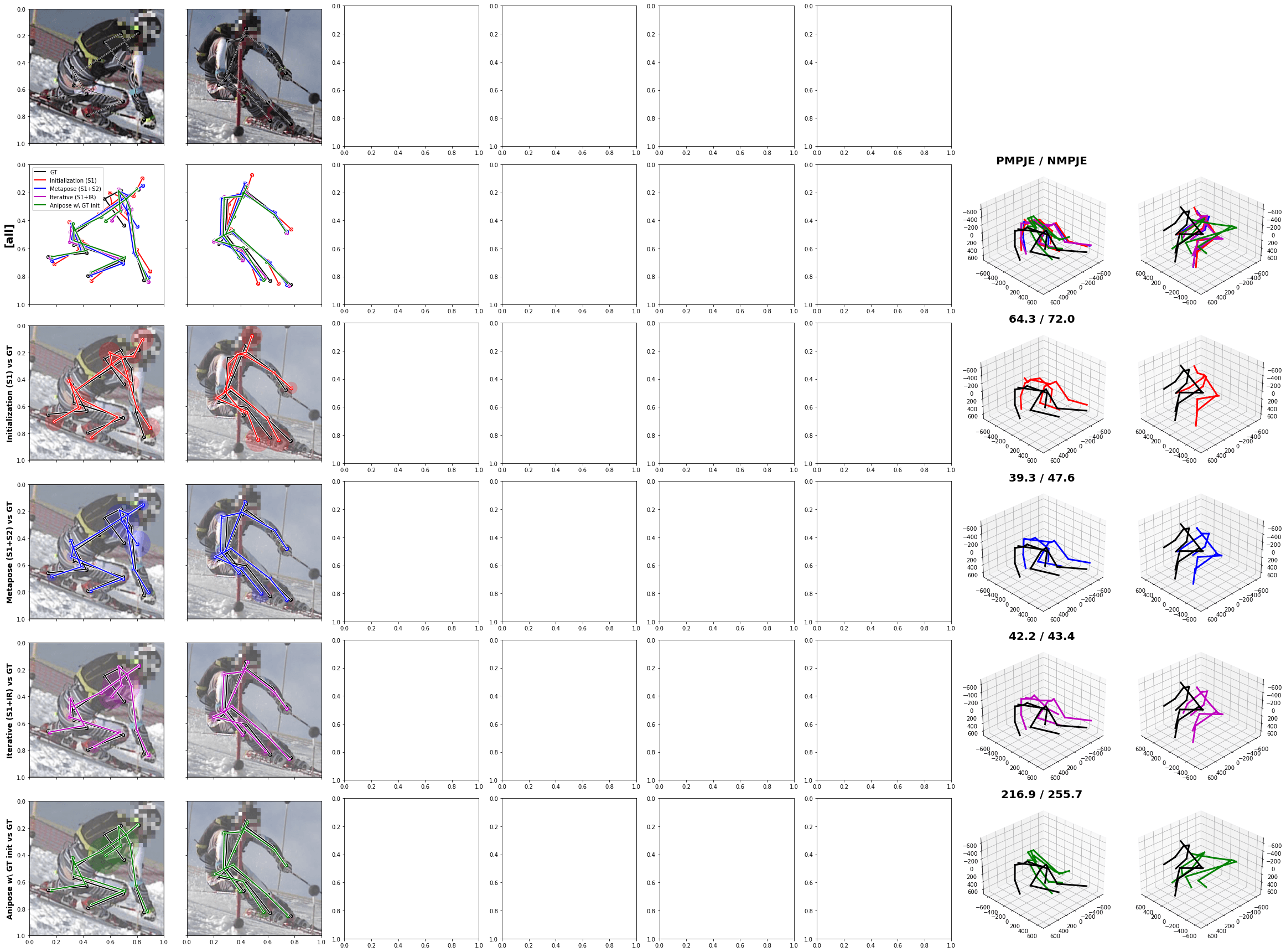}{With \textbf{two cameras} AniPose with GT camera init often yields low reprojection error but bad 3D estimation error}

\exfignewcrop{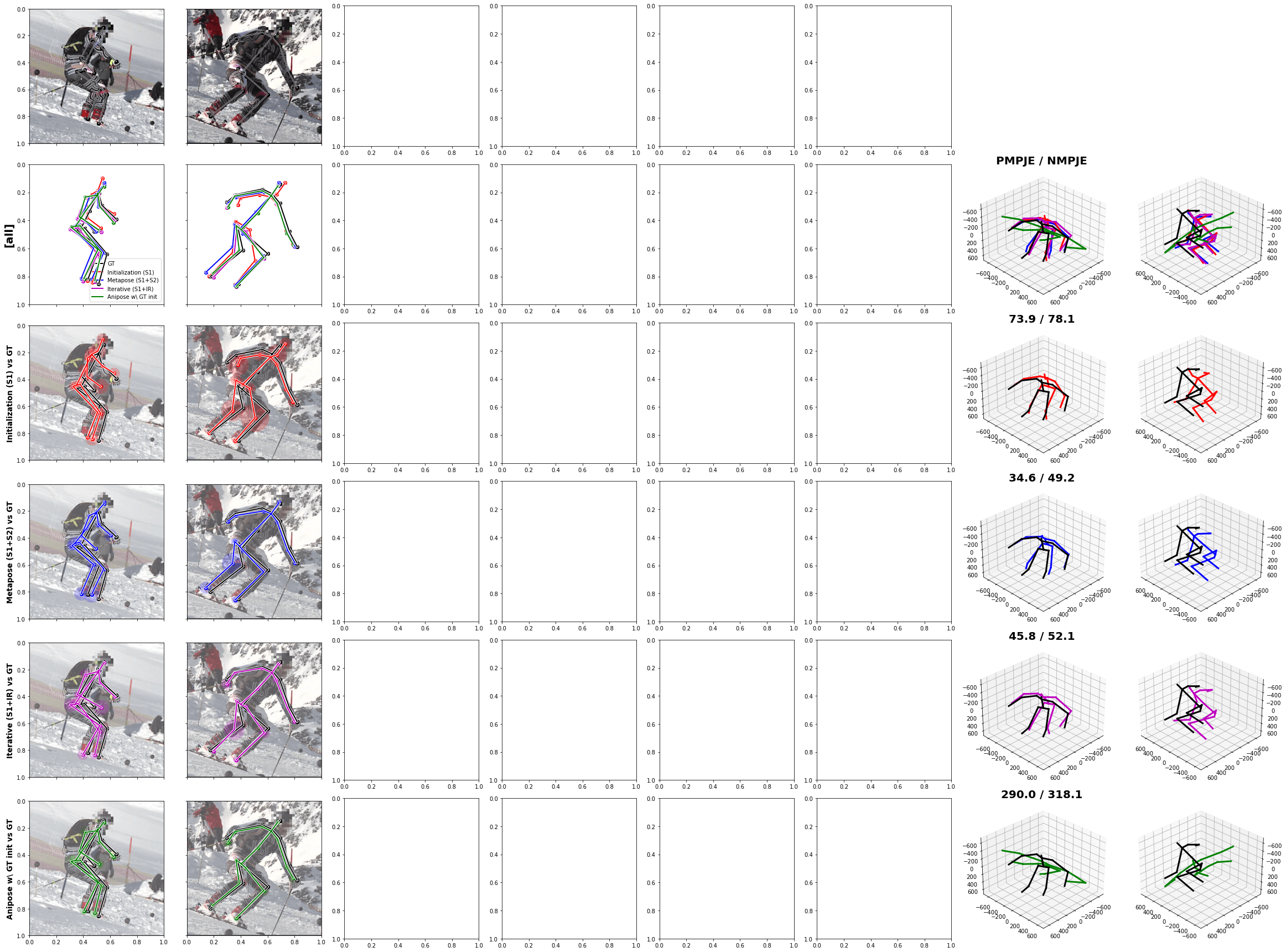}{With \textbf{two cameras} AniPose with GT camera init often yields low reprojection error but bad 3D estimation error}